\documentclass[10pt,journal,compsoc]{IEEEtran}
\usepackage{amsmath,amsfonts}
\usepackage{algorithm}
\usepackage{algorithmicx}
\usepackage{array}
\usepackage[caption=false,font=normalsize,labelfont=sf,textfont=sf]{subfig}
\usepackage{textcomp}
\usepackage{stfloats}
\usepackage{url}
\usepackage{verbatim}
\usepackage{graphicx}
\usepackage[utf8]{inputenc} % allow utf-8 input
\usepackage[T1]{fontenc}    % use 8-bit T1 fonts
\usepackage[breaklinks=true,colorlinks,bookmarks=false]{hyperref}
\usepackage{url}            % simple URL typesetting
\usepackage{booktabs}       % professional-quality tables
\usepackage{amsfonts}       % blackboard math symbols
\usepackage{nicefrac}       % compact symbols for 1/2, etc.
\usepackage{microtype}      % microtypography
\usepackage{amsthm}
\usepackage{amsmath}
\usepackage{graphicx}
\usepackage{amssymb}
\usepackage{diagbox}
\usepackage{multirow}
\usepackage{pifont} % star
\usepackage{bm}
\usepackage{color,xcolor}
\usepackage{caption}
\usepackage{wrapfig}
\usepackage{colortbl}
\usepackage{enumitem}
\usepackage{multirow}
\usepackage{algorithm, algpseudocode}
\usepackage{threeparttable}
\usepackage{mathrsfs}
\usepackage[misc]{ifsym}

\newcommand{\tabincell}[2]{\begin{tabular}{@{}#1@{}}#2\end{tabular}}

\ifCLASSOPTIONcompsoc
  \usepackage[nocompress]{cite}
\else
  \usepackage{cite}
\fi
\ifCLASSINFOpdf
\else
\fi
\begin{document}

\title{Modeling Spatio-temporal Dynamical Systems with Neural Discrete Learning and Levels-of-Experts}
\author{Kun Wang, %~\IEEEmembership{Member,~IEEE,}
        Hao Wu, %~\IEEEmembership{Student Member,~IEEE,}
        Guibin Zhang, %~\IEEEmembership{Member,~IEEE,}
        Junfeng Fang, %~\IEEEmembership{Member,~IEEE,}
        Yuxuan Liang$^{\textrm{\Letter}}$,~\IEEEmembership{Member,~IEEE,}
        Yuankai Wu,~\IEEEmembership{Senior Member,~IEEE,}
        Roger Zimmermann,~\IEEEmembership{Senior Member,~IEEE,}
        and~Yang~Wang$^{\textrm{\Letter}}$,~\IEEEmembership{Senior Member,~IEEE}% <-this % stops a space

\IEEEcompsocitemizethanks{\IEEEcompsocthanksitem K Wang, H. Wu, J.F. fang and Y. Wang are all with University of Science and Technology of China, Hefei, Anhui, P.R.China.
\IEEEcompsocthanksitem Guibin Zhang is with Tongji University and The Hong Kong University of Science and Technology (Guangzhou), P.R.China.
\IEEEcompsocthanksitem Yuxuan Liang is with INTR Thrust \& DSA Thrust, The Hong Kong University of Science and Technology (Guangzhou).
\IEEEcompsocthanksitem Yuankai Wu is with the College of Computer Science, Sichuan University. 
\IEEEcompsocthanksitem Roger Zimmermann is with the University of Singapore, Singapore.
\IEEEcompsocthanksitem ${\textrm{\Letter}}$ Yang Wang and Yuxuan Liang are the corresponding authors. Email: angyan@ustc.edu.cn \& yuxliang@outlook.com.}
}


% \markboth{Journal of \LaTeX\ Class Files,~Vol.~14, No.~8, August~2015}%
% {Shell \MakeLowercase{\textit{et al.}}: Bare Advanced Demo of IEEEtran.cls for IEEE Computer Society Journals}

\markboth{IEEE TRANSACTIONS ON KNOWLEDGE AND DATA ENGINEERING}%
{Wang \MakeLowercase{\textit{et al.}}: Modeling Spatio-temporal Dynamical Systems with Neural Discrete Learning and Levels-of-Experts}



\IEEEtitleabstractindextext{%
\begin{abstract}
% Modeling spatio-temporal dynamics requires estimating states and parameters from a series of observations. Typically, comprehension of these spatio-temporal processes hinges on numerous physical laws. As specialized approaches, dynamical systems, deeply anchored in various physical systems, have been shown to align with the foundational principles of real-world spatio-temporal phenomena.

In this paper, we address the issue of modeling and estimating changes in the state of the spatio-temporal dynamical systems based on a sequence of observations like video frames. Traditional numerical simulation systems depend largely on the initial settings and correctness of the constructed partial differential equations (PDEs). Despite recent efforts yielding significant success in discovering data-driven PDEs with neural networks, the limitations posed by singular scenarios and the absence of local insights prevent them from performing effectively in a broader real-world context. To this end, this paper propose the universal expert module -- that is, optical flow estimation component, to capture the evolution laws of general physical processes in a data-driven fashion. To enhance local insight, we painstakingly design a finer-grained physical pipeline, since local characteristics may be influenced by various internal contextual information, which may contradict the macroscopic properties of the whole system. Further, we harness currently popular neural discrete learning to unveil the underlying important features in its latent space, this process better injects interpretability, which can help us obtain a powerful prior over these discrete random variables. We conduct extensive experiments and ablations to demonstrate that the proposed framework achieves large performance margins, compared with the existing SOTA baselines. 
\end{abstract}

\begin{IEEEkeywords}
Spatio-temporal dynamics, Neural discrete learning, Optical flow estimation.
\end{IEEEkeywords}}

\maketitle

% is not selected <OR> if conference mode is selected - because compsoc
% conference papers position the abstract like regular (non-compsoc)
\IEEEdisplaynontitleabstractindextext

% \fi
\IEEEpeerreviewmaketitle

\ifCLASSOPTIONcompsoc
\IEEEraisesectionheading{\section{Introduction}\label{sec:introduction}}
\else
\section{Introduction}
\label{sec:introduction}
\fi
\IEEEPARstart{T}{he} spatio-temporal physical process is a way of describing a sustained phenomenon involving all objects and matters in the physical world \cite{benacerraf1973mathematical, newell1980physical}. These phenomena range from atoms to the earth, to the far reaches of galaxies and span the entire timeline. Dynamical systems, primarily derived from first principles \cite{pryor2009multiphysics, roberts2022principles, burkle2021deep, yu2020learning, chan2022redunet}, provide a well-established mathematical tool for representing real-world physical processes \cite{anderson1972more}. They model the complex interactions between variables that evolve over time and space, such as Newton's second law \cite{pierson1993corpore}, the advection–diffusion equation \cite{sharan1996mathematical, egan1972numerical}, and the Navier-Stokes equations \cite{shan2006kinetic, mccracken2018artificial}, to name just a few. Formally, dynamical systems involve partial differential equations for analyzing, predicting, and understanding the evolution of the system's state \cite{arrowsmith1992dynamical, hirsch1984dynamical,hirsch2012differential,sideris2013ordinary,verhulst2006nonlinear}.

Traditional spatio-temporal dynamical systems utilize numerical simulations for modeling, and the mathematical philosophies in these systems are universal to have a wide range of applications in classical mechanical systems \cite{wiggins2003introduction, hale2012dynamics, humar2012dynamics}, electrical circuits \cite{wang2020physics, harish2016review}, turbulence \cite{moin1998direct, rogallo1984numerical, orszag1972numerical}, gas \cite{sanyal1999numerical, van2008numerical} and fire science \cite{yang2011experimental, ma2003numerical}, and many other system that evolves with time. Nonetheless, the systems encounter significant challenges when faced with two conditions \cite{bakarji2022discovering, he2021deep}:
(1) \textbf{Uncertainty Principle.} The laws governing complex unfamiliar physical systems are difficult to describe or define intuitively in terms of physical equations, and there are many unknown or partially known equations in nonlinear dynamics \cite{cannon2003dynamics, bezruchko2010extracting, chen2022automated}. (2) \textbf{System equation complexity.} Understanding the dynamics of objectives requires solving PDEs \cite{godunov1959finite, bar1999fitting}. However, with the sharp increase of observable data, the complexity of the system to solve PDEs increases exponentially, leading to delays in the system \cite{yanchuk2017spatio, baker1997mathematical}.

In brief, the performance of numerical simulation methods depends largely on the initial settings and the designed algorithms. Furthermore, it is plagued by the drawbacks of considerable computational resource consumption and the incapacity to meet the demands of real-time prediction.

\begin{figure}[t]
\centering
\includegraphics[width=0.99\columnwidth]{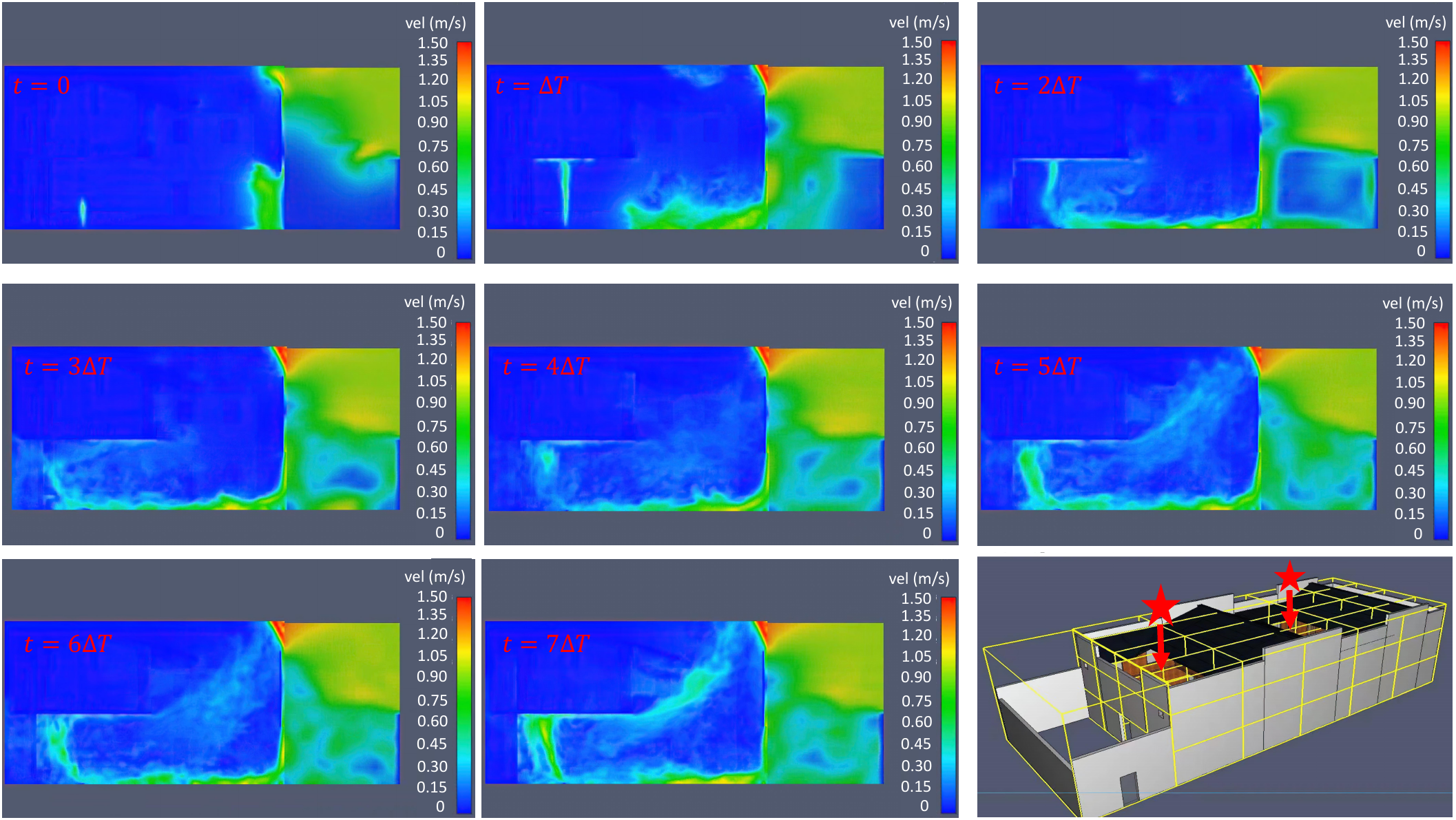}
\caption{An example of dynamic evolution process of fire. Two fires (\textcolor{red}{\text{\ding{72}}}) broke out in a building and we showcase the global and local speed of fire spread.}
\label{fig:fire}
\end{figure}

Recently, Deep Learning (DL) methods have emerged as promising approaches for modeling complex systems with the dilemmas mentioned earlier \cite{graves2012long, mnih2013playing, krizhevsky2017imagenet, creswell2018generative, he2016deep,shi2015convolutional}, as DL provides an efficient optimization framework to automatically, adaptively, and dynamically extract intrinsic patterns from continuous physical processes. Unlike classic dynamic systems, DL frameworks often sacrifice the explicit understandings of physical rules (\textit{aka.,} expert priors), resorting to the large-scale observable data and capturing the implicit patterns that serve as substitutes for physical laws. However, this comes at a cost; discarding expert priors may lead to model collapse under limited data and complex scenarios. Such knowledge of physics ensures that our predictions remain within a testable range without anomalies. To this end, several proposals have integrated dynamic physics laws into their frameworks for facilitating predictions, also known as \emph{Physics-Informed Neural Networks} (PINN).

\noindent \textbf{Research Gap.} In general, a physical process is expressed as a sequence of snapshots, and the modeling objective can be deemed as a Spatio-Temporal (ST) prediction task \cite{dedeep, guen2020disentangling, seo2020physics}. The most existing PINNs achieve the encapsulation of the fundamental physical insights by introducing PDEs into the loss or likelihood functions \cite{raissi2019physics, karniadakis2021physics}. Once the specific expert component (\textit{e.g.}, weather component) is pre-trained for a single scenario, it can be invoked repeatedly with low computational cost. However, a single scenario cannot satisfy various and general physics scenarios, which necessitates a laborious dataset-specific design for prior selection. Worse still, PINNs are commonly employed for addressing challenges related to the state estimation of unknown spatiotemporal points \cite{karniadakis2021physics}. When applied to prediction tasks involving the future evolution of noisy physical processes, PINNs exhibit limitations in terms of compatibility and performance.

\begin{figure*}[!t]
\centering
\includegraphics[width=18cm,height=8cm]{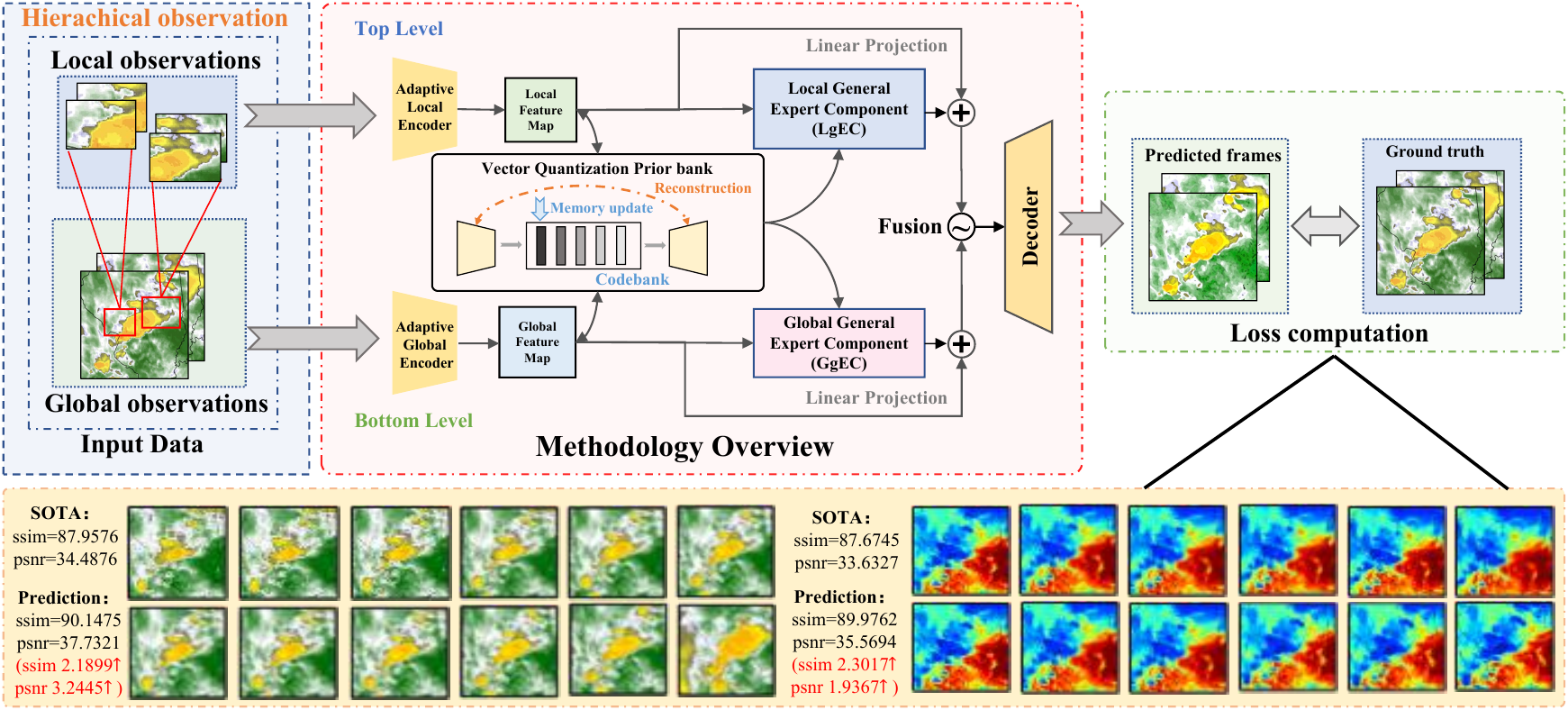}
% \vspace{-1em}
\caption{\textbf{\textit{Top.}} An overview of our framework. \textbf{\textit{Bottom.}} Our proposal framework exhibits exceptional improvements over existing models, demonstrating its factuality and steerability.}
\label{intro}
\vspace{-1em}
\end{figure*}

Furthermore, off-the-shelf ST modules \cite{gao2022earthformer, wang2022predrnn, chen2022automated, DBLP:conf/iclr/PfaffFSB21}, only focus on global understanding, neglecting \emph{local fidelity} -- local dynamics may differ from global dynamics. For instance, Fig. \ref{fig:fire} shows velocity slices of two successive frames illustrating fire's progression. Globally, the fire's flow rate is continuously increasing as it intensifies and expands. However, at a local level, the convection caused by multiple indoor fire points can lead to some number of areas experiencing lower flow rates due to mutual interference. \textit{We argue that this phenomenon is prevalent, with local characteristics manifesting in specific spatiotemporal scenes, while the global situation is entirely opposite.  It is of immense practical significance to pay close attention to these local areas, especially in emergency situations such as fires, where knowledge of low flow rates can better guide the crowd and assist firefighters in dealing with the crisis more efficiently.} To address the above issue, it is necessary to develop custom models or modify existing ones to better capture the local dynamics of the systems.

Towards this end, we introduce a general expert-level physics modeling scheme that operates at multiple granularities with good interpretability, dubbed \textbf{PhysicNet}. As depicted in Fig. \ref{intro},  we commence by hierarchically observing various receptive fields to globally and locally enhance the observed data. Then we send the hierarchical observations to two pipelines; in the top-level and bottom-level pipelines, we harness adaptive global/local encoders, respectively, and automatically characterize the state of physical processes from multiple viewpoints. We utilize a vector quantization (VQ) prior bank, referred to as ``codebank'' to identify the most compact, comprehensive, and discrete embedding space and priors, instead of allocating capacity towards noise and imperceptible details. By combining VQ \cite{van2017neural, fortuin2018som} with the representation capabilities of neural networks, our model can enhance its expressive ability and improve its local insight and fidelity.

After an iterative process of transferring and transforming, we transmit the feature map, which contains physical knowledge, to the global/local general expert component (called GgEC or LgEC), enabling them to perceive physical processes. These expert components are utilized to process data that conforms to a continuous physical context process. Concretely, we introduce an optical flow estimation constraint term to describe the direction and magnitude of motion for each pixel in a sequence of observations, which ensures that its process conforms to the evolution process of the physical situation. Our work is simple yet effective, allowing for accurate predictions of local details. We are confident that the insights gained from this study will inspire further research on modeling physical processes and their potential applications. In short, our contributions can be summarized as:

% \section*{Acknowledgments}
% This should be a simple paragraph before the References to thank those individuals and institutions who have supported your work on this article.

\begin{itemize}[leftmargin=*]
   \item We present a data-driven neural network that integrates expert knowledge and extracts crucial underlying patterns for accurate observations while maintaining interpretability. Furthermore, by emphasizing the integration of local characteristics into the model,  we are proposing a new direction with ongoing and future efforts in the field. Since our framework employs a data-driven approach to capture crucial information from noisy datasets, it mitigates the data limitations inherent in PINNs. Consequently, our approach exhibits enhanced robustness in real-world scenarios, thereby showcasing greater resilience.

   \item Our paradigm can automatically discover physical evolution laws from high-dimensional data by combining Vector discrete learning theory and neural network representations. We refine and extract hidden state variables from the network encoding, enabling powerful and interpretable dynamical system prediction in discrete, sparse parsimonious dynamics.
   % \vspace{-0.1em}

   \item Our optical flow estimation module effectively captures the spatio-temporal data variations, ensuring that the model accurately grasps ST dynamics, which in turn guarantees the stability of predictions.
   
   \item Extensive experiments on four real-world datasets across various baselines to demonstrate the effectiveness and efficiency of the proposed framework. We highlight some shocking performances of our method on SEVIR-Infrared Satellite imagery dataset in the bottom part of the Fig. \ref{intro}.

   % \item We believe our work serves as an important precursor for future research. By emphasizing the integration of local characteristics into the model, we are proposing a new direction that we believe will resonate with ongoing and future efforts in the field. The attention to local details, while maintaining a global perspective, offers a nuanced approach that has the potential to refine and even transform the way models are constructed and predictions are made.

\end{itemize}

\section{Preliminary \& Related work} \label{related work}
\textbf{Problem Formulation.} Physical systems evolve continuously based on their dynamics, which can be represented as a function with environment space $X$ and state space $S\subset X$: $X_{t+\Delta t} = \mathcal{F}\left( X_{t} \right)$, $\mathcal{F}$ describes system state evolution from current to next time step. We address the forecasting of physical dynamics as a spatiotemporal forecasting problem, \emph{i.e.}, mapping high-dimensional visual observations $\left[\mathcal{X}_t\right]_{t=1}^T, \mathcal{X}_t \in \mathbb{R}^{C \times H \times W}$ to a relatively low-dimensional embedding $L_s$, which is then projected onto a spatiotemporal sequence of $K$ future steps $\left[\mathcal{X}_{T+t'}\right]_{t'=1}^K,\mathcal{X}_{T+t'} \in \mathbb{R}^{C \times H \times W}$, where $H$ and $W$ denote the number of spatial grids with $C$-dimensional observations on each grid. 

\noindent \textbf{Local and Global Insight.} We introduce, for the first time, a combined global and local spatiotemporal modeling approach that aims to simultaneously capture both overall trends and localized intricacies. Drawing inspiration from related work in contrastive learning \cite{caron2021emerging}, we define local insights as coverage areas that are below 50\% of the original region. Given the unknown nature of crucial local regions, we adopt a random cropping method to extract $\cal H$ distinct local perspectives. A larger value of $\cal H$ necessitates greater computational resources, whereas a smaller $\cal H$ may result in the loss of significant local areas.

\noindent \textbf{Physics-Informed Neural Networks (PINN).} Scientists explore the potential of PINNs in modeling system dynamics by incorporating physics priors or logical constraints into neural networks \cite{stewart2017label, watters2017visual, kashinath2021physics, wang2020towards, xue2020amortized, sanchez2018graph, wu2017learning, raissi2019physics, lu2019deeponet}. Physical control equations connect deep neural networks to numerical differential equations, resulting in the development of generalized architectures with specific functionalities \cite{sirignano2018dgm, han2018solving, meng2020ppinn} that reduce complexity in simulating physical processes. Our research aims to model the entire process of physical dynamics based on PINNs, effectively capturing non-linear and dynamic relationships among system components \cite{liu2022predicting, rao2021hard, rao2022discovering, huang2022meta}. Unfortunately, PINNs are typically applied to idealized noise-free data for addressing state estimation of unknown spatiotemporal points. They do not exhibit significant expressive power when applied to our real-world spatiotemporal prediction problems.

\noindent \textbf{Optical flow estimation.} Substantial advancements in optical flow estimation have been achieved with the incorporation of deep learning \cite{horn1981determining, zach2007duality, brox2010large}.  Traditional methods relied on variational principles \cite{menze2015discrete}, optimizing flow prediction through continuous optimization with various objective terms, which has led to discrete optimization improvements for modeling optical flows \cite{chen2016full}.  More recent efforts have pivoted towards deep learning frameworks that employ coarse-to-fine strategies and iterative refinements.  Notably, the RAFT model emerged as a pivotal development, introducing a convolutional GRU update mechanism and achieving state-of-the-art performance \cite{teed2020raft}.  Enhancements to this model have been made by \cite{jiang2021learning}, who integrated a self-attention-style global motion aggregation module to refine the handling of occlusions.  In our work, we first involve the optical flow in to ST prediction task to constrain the prediction process to be stable and reliable.

\noindent \textbf{Neural Discrete Learning.} Recently, models which leverage discrete representation through vector quantization have been proposed. Milestones such as VQ-VAE \cite{van2017neural} and VQ-VAE2 \cite{razavi2019generating} show great promise in image, speech and video generation, leading to numerous follow-up studies \cite{walker2021predicting, bocus2023streamlining, liu2021conditional, zhao2018unsupervised}. Broadly speaking, VQ-VAE consists of an encoder network, a quantization module, a decoder network, and a loss function. The encoder network takes an input image and encodes it into a low-dimensional latent representation. The quantization module then maps this continuous latent representation to a discrete codebank of embedding vectors. The decoder network takes the discrete code and decodes it back into a reconstructed image. Our work is the first to incorporate vector quantization encoding techniques into physical processes, for learning important latent features in both global and local physical states, and aiding in better priors for temporal processes.

\noindent \textbf{Spatio-temporal prediction methods.} Spatio-temporal data mining, a merger of classical spatial data mining and temporal data mining, focuses on the extraction of patterns, relationships, and knowledge from spatial and temporal data. Generally, spatio-temporal prediction methods can be classified into CNN-based approaches \cite{oh2015action, mathieu2015deep, tulyakov2018mocogan}, RNN-based techniques \cite{ranzato2014video, srivastava2015unsupervised, villegas2017learning, villegas2018hierarchical, kim2019variational, wang2022predrnn}, and other models encompassing combinations \cite{weissenborn2019scaling, kumar2019videoflow}, as well as transformer-based models \cite{dosovitskiy2020image, gao2022earthformer, bai2022rainformer}. Interestingly, while various existing models built upon graph neural networks (GNNs) are available, their primary focus revolves around graph data handling \cite{sun2020predicting, wang2020deep, jiang2021dl, jiang2019censnet, wang2022a2djp}, which falls outside the purview of our study.

\noindent \textbf{Deep Learning for Dynamical Systems.} DL has become a popular tool for predicting dynamical system behavior. As an early study in this area, ConvLSTM integrates CNNs with LSTM to jointly capture spatial and temporal information in precipitation data \cite{shi2015convolutional}. Advanced models, such as PredRNN \cite{wang2017predrnn}, E3D-LSTM \cite{wang2019eidetic}, and SimVP \cite{gao2022simvp}, further improve accuracy and robustness by devising sophisticated architectures. \cite{guen2020disentangling} presents a recursive physical unit that separates PDE dynamics from unknown complementary information to enhance predictions using DL. Researchers have also investigated modeling system dynamics from video data, proposing various methods for extracting physical parameters from input data \cite{chen2022automated, yang2022learning, jaques2019physics}. However, the generalization ability of these models is limited by the requirement of known equations governing system dynamics.

\section{Methodology}

\begin{figure*}[!t]
\centering
\includegraphics[width=18cm,height=6cm]{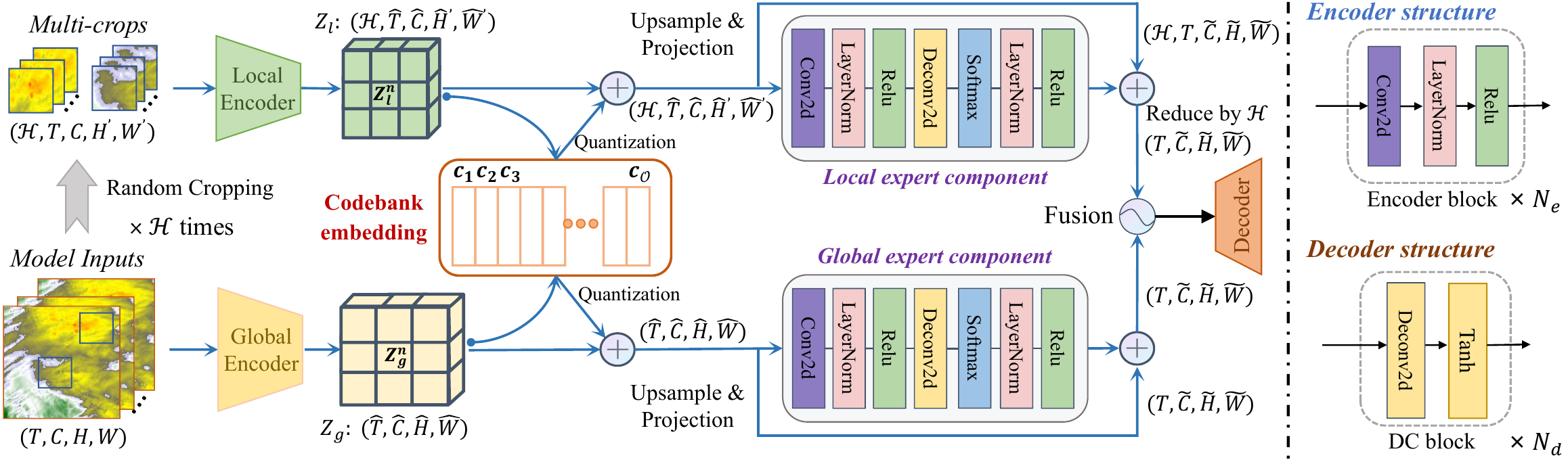}
\caption{The pipeline of our model. Conv2d: 2D convolutional layer. Deconv2d: deconvolutional operator for upsamling. DC block: Deconvolutional block.}
\label{method}
% \vspace{-1.5em}
\end{figure*}

In this paper, we encapsulate physical process prediction as a spatiotemporal forecasting task as \cite{gao2022earthformer} does. The pipeline of our PhysicNet is presented in Fig. \ref{method}, whose input is the historical observations, i.e., a spatiotemporal sequence $\left[\mathcal{X}_t\right]_{t=1}^T, \mathcal{X}_t \in \mathbb{R}^{C \times H \times W}$ at the left bottom.  We aim to parallelly predict the future $K$ steps of the spatiotemporal sequence. In the following parts, we will delineate our model components by introducing Fig. \ref{method} from left to right.

\subsection{Random Cropping to Multi-Grained Observations} \label{method_1}
The observations used in this study, which include global and local views, are encoded using similar forms (but with different kernel sizes). However, many current models \cite{gao2022simvp, yu2020efficient, wang2017predrnn} mainly focus on global physical processes, which can result in inherent limitations: \textit{global insight can compromise local fidelity since globally significant information may not be important or may not even appear in the local context. Consequently, such models may fail to provide reliable understanding of specific instances.} Going beyond the global view, several (in this paper, we set ${\cal H}=3$ local views for the trade-off between the accuracy and the efficiency, the left side of Fig \ref{method}) local views of resolution ${32^2}$ covering only small areas (less than $50\%$) of the original image. Similar to \cite{caron2021emerging}, we randomly select the crops in original input, since we do not know at first which local observations have specific physical processes. For ease of understanding, we characterize the global observation contains $T$, $C$, $H$, and $W$ dimensions of the entire sequence while local views with smaller height $H'$ and width $W'$. In this fashion, we explicitly embody the local importance by inspecting the local areas.

\subsection{Adaptive Encoders Towards Global/Local Expressivity}

After getting multi-grained observations, we employ adaptive encoders toward acquiring embedding representations for information extraction. Let the tensor ${{\cal X}_g} \in {\mathbb{R}^{{\cal H} \times T \times H \times W}}$ denote global observation and ${\cal X}_{l} \in \mathbb{R}^{{\cal{H}} \times T \times C  \times H^{'} \times W^{'}}$ represent the fine-grained local views, adaptive local and global encoder stack $N_e$ blocks (Conv2d+LayerNorm+Relu) for extracting spatial features, \textit{i.e.,} convoluting $C$ on $(H, W)$. The hidden feature of global embedding ($Z_g$) and local embedding ($Z_l$) can be written as: 

\begin{equation}\small
\begin{aligned}
\label{eq:embedding}
& \text{Global:} \; Z_{g} = {\left\{ \operatorname{Relu}\left( \operatorname{LayerNorm}\left( \operatorname{Conv2d}\left( \mathcal{X}_{g} \right) \right) \right) \right\}|}_{\times N_{e}}; \\
& \text{Local:} \; Z_{l} = {\left\{ \operatorname{Relu}\left( \operatorname{LayerNorm}\left( \operatorname{Conv2d}\left( \mathcal{X}_{l} \right) \right) \right) \right\}|}_{\times N_{e}},
\end{aligned}
\end{equation}

\noindent where the inputs are ${{\cal X}_g}$ / ${{\cal X}_l}$ and the output are ${Z_g} \in {\mathbb{R}^{{\hat T} \times {\hat C} \times {\hat H} \times {\hat W}}}$ and ${Z_l} \in {\mathbb{R}^{ {\cal H} \times {\hat T} \times {\hat C} \times {\hat H}^{'} \times {\hat W}^{'}}}$, respectively. $(  \cdot  ){|_{ \times {N_e}}}$ indicates that the mapping blocks have been applied $N_e$ times. In our work, we set the local encoder kernel size to $3 \times 3$ and the global encoder kernel size to $7 \times 7$.

\begin{figure}[h]
\centering  
\includegraphics[width=0.70\columnwidth]{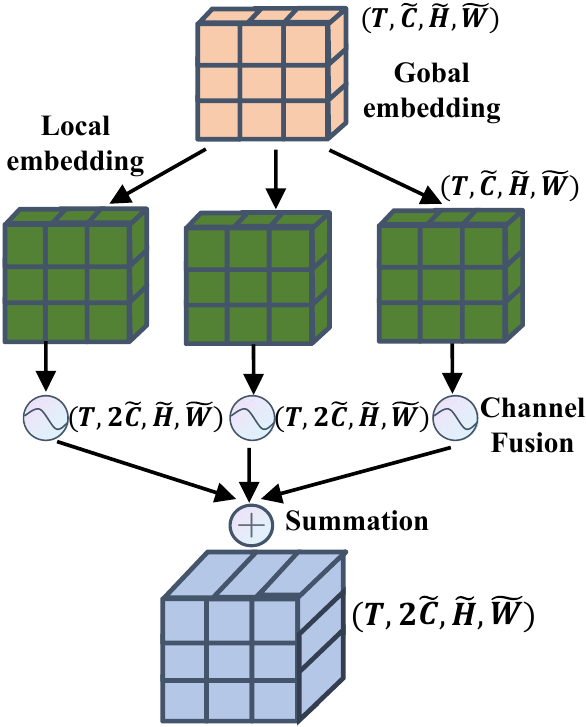}
% \vspace{-0.6em}
\caption{Fusion process of global and local pipelines. }\label{fusion}
\label{ablation}
\end{figure}

Since there are multiple crops in the local pipeline, we conduct feedforward process three times of local views but only one time of global view. After characterizing the global view as a feature map before the decoder, we perform local transformation three times. Finally, we conduct three times channel fusion between global and local embeddings and then process the summation operator. It's worth emphasizing that during the forward propagation process, we only calculate the global input once, and do not repeatedly calculate its update. Once the global data is computed, we wait for the updated vector of the local view to fuse, which significantly reduces computing power expenditure (process is shown in Fig. \ref{fusion}, for ease of understanding, we omitted the T dimension from the presentation).

\subsection{Vector Quantization Prior bank}

Recently, neural discrete representation learning \cite{van2017neural, fortuin2018som, pervez2021stability, liu2021cross} has shown great prominence in generative modeling of image \cite{van2017neural, razavi2019generating, peng2021generating}, video \cite{yan2021videogpt, walker2021predicting, zhang2020videogen} and audio \cite{garbacea2019low, tjandra2020transformer}, to name just a few. Unlike continuous coding models, neural discrete learning involves discrete codes with a prior distribution. Drawing inspiration from this, we first involve this technology in the realm of modeling physical processes, dubbed as \textit{vector quantization prior bank or codebank}, for highlighting the important features in latent space rather than noise and imperceptible details.  

\noindent \textbf{Vector quantization prior bank.} We define that the latent embedding space of codebank $c = \left\{ c_{1},c_{2}\ldots c_{\cal{O}} \right\} \in \mathbb{R}^{\cal{O} \times D}$, where $\cal{O}$ is the size of the discrete latent space (i.e., $\cal{O}$-way categorical) and $D$ is the dimensionality of each quantized vector. As shown in Fig. \ref{method}, encoder produces outputs $Z_g$ and $Z_l$, the discrete latent variables are obtained through a nearest neighbor look-up in the shared embedding space as demonstrated in Eq. \ref{vq1}. Here we define the posterior categorical distribution ${q_i}\left( Z \right)$ as the one-hot form:

\begin{equation} \small
\label{vq1}
q_i(Z) = \begin{cases}
1, & i = \arg \mathop {\min }\limits_j \| Z - c_j \|_2; \;\; Z=Z_l \;\; or \;\; Z_g \\
0, & \text{otherwise}
\end{cases}
\end{equation}

The input to general expert component is the corresponding latent vector $c_i$ as shown in Eq. \ref{vq2}:
% \vspace{-0.4em}
\begin{equation}\small
\begin{aligned}
\label{vq2}
    & {Z^{\left( {VQ} \right)}} = {c_j}\;\; \text{where} \;i = \arg\min_j \| Z - c_j \|_2; \\
    & {Z^{\left( {VQ} \right)}} = Z_g^{\left( {VQ} \right)}\;or\;Z_l^{\left( {VQ} \right)},
\end{aligned}
\end{equation}
where $Z_g^{\left( {VQ} \right)}$ and $Z_l^{\left( {VQ} \right)}$ are the global and local quantized vectors from the codebank, respectively. It's worth emphasizing that there is no real gradient propagation here. To this end, we adopt the reparameterization trick \cite{van2017neural} and just copy gradient from expert component input to encoder output. Afterward, we combine (sum) the outputs from the codebank and the encoder to facilitate information fusion. We proceed to consider the loss of our codebank part, here we follow \cite{van2017neural} to give our training objective as:  $\mathcal{L}^{(VQ)} = \left| \middle| \text{sg}\lbrack Z\rbrack - c \middle| \right|_{2}^{2} + \beta\left| \middle| Z - \text{sg}\lbrack c\rbrack| \right|_{2}^{2}$, here sg denotes the stopgradient operator that is defined as identity at forward computation time and has zero partial derivatives. Here, we have removed the reconstruction loss as our output does not directly feed to the decoder for reconstruction task, but rather is sent to the expert component.  Here existing two loss function ${\cal L}_g^{( {VQ})}$ (global loss) and ${\cal L}_l^{( {VQ} )}$ (local loss), they share the same forms as shown above. \textit{Since the physical process represents a sustained phenomenon that can be better comprehended through previous observation, our code bank can learn reliable priors and continuous physical events, rather than static ones.}

\noindent \textbf{Correlation with physical process.} As illustrated above, our codebank can acquire a dependable prior rather than a static one; physical processes are ongoing phenomena that can be comprehended through past observations. Additionally, our vector quantization strategy demonstrates proficiency in identifying discrete events that define these physical processes.

\subsection{Expert Component \& Unified Decoder for Prediction}

Optical flow field (also called motion field) refers to the pattern of apparent motion of objects in a visual scene. An optical flow field is a two-dimensional vector field that describes the direction and magnitude of motion for each pixel in an image or sequence of images \cite{horn1981determining}. In our work, the local and global expert modules process image sequences of varying shapes and compute their corresponding optical flow field in natural physical processes \cite{de2019deep, brox2004high, bruhn2005lucas, horn1981determining}. Specifically, the local/global expert modules take the embeddings with dimensions of $( {\hat T,\hat C,{{\hat H}^{\rm{'}}},{{\hat W}^{\rm{'}}}} )$ and $( {\hat T,\hat C,\hat H,\hat W} )$, respectively. The expert modules process embedding sequences and output optical flow fields (Fig. \ref{method}). Convolutional layers are utilized to extract features (local: $3 \times 3$, global: $7 \times 7$), while deconvolution layers, also referred to as transposed convolutional layers, are employed to upsample feature maps and generate precise optical flow fields (also ensuring projecting onto the target dimension). In some optical flow methods, deconvolution layers are referred to as ``optical flow decoders'' \cite{dosovitskiy2015flownet, fischer2015flownet, ilg2017flownet}. 

In our implementation, the small area pixel motion field can capture more local details, while the global motion field has a larger receptive field and is capable of focusing on a wide pixel area. Expert components use CNN-based frameworks \footnote{During this process, several commonly-used CNN architectures, such as U-Net, FlowNet, can be employed. These architectures comprise multiple Conv2d and Deconv2d, as well as other types such as pooling, activation.} for extracting features and converting features into high-level semantic representations for optical flow estimation. Generally, In general, the optical flow estimation model takes the concatenation of two time-series feature maps as input (assuming $(2c', h', w')$ dimension) and outputs optical flow field $(u, v)$ in dimension $(2, h_{of}, w_{of})$ \cite{ilg2017flownet}. In this work, we first map the outputs of the expert modules to a high-dimensional semantic representation of size $( {T,\tilde C,\tilde H,\tilde W})$ to mimic optical flow estimation. Then, we also perform linear projection on the inputs before the expert modules, mapping them to the same high-dimensional representation $( {T,\tilde C,\tilde H,\tilde W} )$. For simplicity, we use convolutional layer (Conv2d) for the linear projection, and more detailed information of linear projection can be found in the appendix. We then match the outputs of our analogical optical flow estimation and linear projection, and introduce an additional optical flow estimation loss to enforce smoothness during the matching phase: 
% \vspace{-0.4em}
\begin{equation} \footnotesize
    {{\cal L}_{of}} = \mathop \sum \nolimits_t^T \mathop \sum \nolimits_{i,j} \left| {{I_t}\left( {i,j} \right) - {I_{t + 1}}\left( {i + {u_t}\left( {i,j} \right),j + {v_t}\left( {i,j} \right)} \right)} \right|,
\end{equation}
where ${{\cal L}_{of}}$ denotes optical flow estimation loss, ${I_t}( {i,j})$ stands for channel tensor of high-level semantics at $t$ point \cite{shi2015convolutional}. ${u_t}\left( {i,j} \right)$ and ${v_t}\left( {i,j} \right)$ represent displacement vectors from time $t$ to time $t+1$ point. Unlike traditional models for optical flow estimation, our operations focus on high-level semantics, and the $T$-dimension may not exhibit clear temporal characteristics.

\textbf{Correlation with physical process.} Traditional ST frameworks do not have a direct constraint on the smoothness of process. However, in this study, we introduce supervision from the perspective of the motion vector field, which can be considered a universal operator applicable to a wider range of physical process tasks. By introducing the concept of the motion field, our model can easily model the dynamical systems in a powerful and interpretable way.

\vspace{-0.7em}
\subsection{Decoder}
\textbf{Fusing local and global representations.} As there are multiple crops in the local pipeline, we conduct feedforward process three times of local views but only one time of global view. As shown in Fig. \ref{method}, we reduce the global representation on the first dimension using a summation operation. Subsequently, we execute channel fusion between global and local embeddings. 

\noindent \textbf{Decoder and optimization.} In our model, we use only three deconvolutional layers  for the final decoding (with a DC block and $N_d$=3), which we find to be both simpler and more efficient, without the need to introduce additional parameters. We optimize our framework by adopting loss function ${\cal L} = {{\cal L}_{of}} + {{\cal L}^{\left( {VQ} \right)}} + {{\cal L}_{mse}}$, where ${\cal L}_{of}$, ${\cal L}^{\left( {VQ} \right)}$ and ${\cal L}_{mse}$ denote optical flow, codebank and MSE loss, respectively.

% {\appendices
% \section*{Proof of the First Zonklar Equation}

% \begin{figure}[t]
% \centering
% \includegraphics[width=0.99\columnwidth]{appa.pdf}
% \caption{An example of dynamic evolution process of fire.}
% \label{fig:fire}
% \end{figure}
% In this section, we show an example of fire evolution to help better understand global and local differences. As shown in Fig. \ref{fig:fire}, globally, the fire's flow rate is continuously increasing as it intensifies and expands. However, at a local level, the convection caused by multiple indoor fire points can lead to some number of areas experiencing lower flow rates due to mutual interference. \textit{We firmly demonstrate that this phenomenon is prevalent, with local characteristics manifesting in specific spatiotemporal scenes, while the global situation is entirely opposite.  It is of immense practical significance to pay close attention to these local areas, especially in emergency situations such as fires, where knowledge of low flow rates can better guide the crowd and assist firefighters in dealing with the crisis more efficiently.}

\section{Experiments}\label{sec:exp}
In this section, we experimentally assess, analyze, and rationalize the impact of our framework on three real-world physical datasets and a synthetic dataset. We follow some previous endeavors \cite{de2019deep, guen2020disentangling, wang2022predrnn, wu2021motionrnn, gao2022earthformer, chen2022automated} to unitize three metrics: 1) \textit{Structural Similarity Index Measure (SSIM)}: SSIM is a metric for measuring the similarity between two images. 2) \textit{Peak Signal-to-Noise Ratio (PSNR)}: PSNR is a measure of video or image quality that compares the original signal to the compressed or transmitted signal. 3) \textit{MSE (Mean Squared Error)}: MSE calculates the average of the squared differences between the predicted and actual values.  \textit{In our experiments, we consistently use the previous 10 frames of observational data to predict the subsequent 10 frame for base settings.}

\begin{table}[htbp]\footnotesize
  \centering
  \setlength{\tabcolsep}{3.2pt}
  \caption{Statistics of the datasets used in the our paper.}
  % \vspace{-0.3em}
    \begin{tabular}{ccccccc}
    \toprule
    \multirow{2}[4]{*}{Dataset} & \multicolumn{3}{c}{Size} & Seq.  & Len.  & Spatial Resolution \\
\cmidrule{2-7}          & train  & val   & test  & in    & out   & $H \times W$ \\
    \midrule
    Reaction–diffusion & 1500  & 500   & 500   & 10     & 10     & $128 \times 128$ \\
    Fire  & 1500  & 500   & 500   & 10     & 10     & $128 \times 128$ \\
    SEVIR & 4500  & 500   & 1000  & 10    & 10    & $384 \times 384$ \\
    MovingMNIST & 1500  & 500   & 500   & 10    & 10    & $128 \times 128$ \\
    \bottomrule
    \end{tabular}%
  \label{tab:dataset}%
\end{table}%
% \vspace{-0.3em}

\noindent \textbf{Datasets \& Main Baselines.} We conduct experiments on three spatio-temporal physical datasets (Reaction–diffusion \cite{chen2022automated}, Fire \cite{chen2022automated}, SEVIR \cite{veillette2020sevir}) and a synthetic dataset (MovingMNIST \cite{srivastava2015unsupervised}), to verify the effectiveness of our approach. Except for SEVIR (4500 images for training, 500, 1000 for validation and testing), we chose 80\%, 10\%, and 10\% of the raw data as the training, validation, and test datasets, respectively. We compare our framework with state-of-the-art physical constraint models (I, II, VI) and video understanding methods (III, IV, V): 

\begin{itemize}
    \item (I) \textbf{Ad-fusion} \cite{de2019deep} demonstrates how to use physical principles to design a neural network for complex prediction tasks, specifically for transport problems that follow advection-diffusion principles. 

    \item (II) \textbf{PhyDNet} \cite{guen2020disentangling} has two branches and separates PDE dynamics from unknown complementary information. It also introduces a new recurrent physical cell for PDE-constrained prediction in latent space, inspired by data assimilation techniques. 

    \item (III) \textbf{PredRNN-V2} \cite{wang2022predrnn} possesses a zigzag memory flow that goes up and down all layers, allowing visual dynamics to communicate and adding to the original memory cell of LSTM. 

    \item (IV) \textbf{MotionRNN} \cite{wu2021motionrnn} proposes a new RNN unit called MotionGRU, which can transient variation and motion trend in a unified way. 

    \item (V) \textbf{Earthformer} \cite{gao2022earthformer} is a Transformer-based architecture, with a novel module termed \textit{Cuboid Attention} mechanism to efficiently learn spatiotemporal representations. 

    \item (VI) \textbf{MathRel} \cite{chen2022automated} presents a principle to determine the probable number of state variables in an observed system and their potential attributes. 
\end{itemize}

\noindent \textbf{Frameworks specifically for PDEs.} To comprehensively validate the generalization ability, we also conducted additional tests using some models specifically designed for addressing continuous and PDE problems. These models include Fourier Neural Operator (FNO) \cite{li2020fourier}, Adaptive Fourier Neural Operator (AFNO) \cite{guibas2021adaptive}, and TF-Net \cite{wang2020towards}. All three models are tailored to handle continuous properties, especially PDEs.

\noindent \textbf{Frameworks for video understanding.} We also experimented with some video understanding baselines, such as MIM \cite{wang2019memory} (CVPR 2019), CrevNet \cite{yu2020efficient} (ICLR 2020) and SimVP \cite{gao2022simvp} (CVPR 2022). Since our work can also be interpreted as a video frame prediction task, we aim to compare these frameworks to demonstrate the scalability of our algorithm.

\begin{table}[htbp]\footnotesize
  \centering
    \setlength{\tabcolsep}{7.0pt}
  \caption{Implementation details on four datasets.}
  % \vspace{-0.5em}
    \begin{tabular}{cccccccc}
    \toprule
    Dataset & learning rate & optimizer & $\beta$     & epoch  \\
    \midrule
    Reaction–diffusion & 0.01  & SGD   & 0.99  & 1000   \\
    Fire  & 0.01  & SGD   & 0.99  & 1000   \\
    SEVIR & 0.01  & SGD   & 0.99  & 20000  \\
    MovingMNIST & 0.01  & SGD   & 0.99  & 500   \\
    \bottomrule
    \end{tabular}%
  \label{tab:hyper}%
\end{table}%

\noindent \textbf{Experimental settings.} We train our framework and the state-of-the-art methods (SOTAs) with the same experimental settings, including learning rate, optimizer, \textit{etc}. For ease of understanding, we summarize detail in Tab. \ref{tab:hyper}. In our paper, we generate \textbf{encoder configurations} (ours-$S/B/L$) by adjusting $N_{e}=2,4,8$. Concretely, small ($S$), base ($B$) and large ($L$) models have $2$, $4$ and $8$ blocks, respectively. We also test the configuration of the decoder, including the depth ($2,3,4,5,6$) and form of the blocks (DC, CL+DC, CL+DC+R)\footnote{Here we abbreviate convolutional, deconvolutional and residual layer as CL, DC and R, respectively.}. However, we find that use three DC blocks can yield the best performance. Specific experimental results will be presented later in this paper. It is important to note that all results discussed herein are based on 3 DC block and $256 \times 64$ codebank size  without particular emphasis. All experiments are conducted on a server with 4 V100 GPUs using a batch size of 64.

\begin{table*}[!t]\footnotesize
\setlength{\tabcolsep}{7.0pt}
  \caption{Comparison with the state-of-the-art methods on MSE, SSIM and PSNR. We report the mean and standard deviation of the results from three experimental runs.} % Metrics are scaled to be in a similar range across datasets to ease comparison, and 
  % \vspace{-0.5em}
  \label{tab:main}
  \centering
  \begin{threeparttable}
  \begin{tabular}{c|c|ccccccc}
    \toprule
    % & \multirow{2}{*}{Metrics} & \multicolumn{7}{c}{Baselines} \\
    & Metrics & Ad-fusion & PhyDNet & PredRNN-V2 & MotionRNN  & Earthformer & MathRel  & PhysicNet (ours) \\
    \midrule
    \tabincell{c}{Reaction \\ -diffusion} & \tabincell{c}{MSE ($ \downarrow $) \\ SSIM ($ \uparrow $) \\ PSNR ($ \uparrow $)} & \tabincell{c}{${30.04_{\pm 0.45}}$ \\ ${0.866_{\pm 0.002}}$ \\ ${29.23_{\pm 0.23}}$} & \tabincell{c}{${9.77_{\pm 0.28}}$ \\ ${0.837_{\pm 0.039}}$ \\ ${27.25_{\pm 0.37}}$} & \tabincell{c}{${29.60_{\pm 0.39}}$ \\ ${0.913_{\pm 0.019}}$ \\ ${29.02_{\pm 0.02}}$} &  \tabincell{c}{${6.69_{\pm 0.34}}$ \\ ${0.866_{\pm 0.024}}$ \\ ${28.97_{\pm 0.02}}$} & \tabincell{c}{${6.61_{\pm 0.17}}$ \\ ${0.864_{\pm 0.015}}$ \\ ${28.47_{\pm 0.04}}$} & \tabincell{c}{${6.44_{\pm 0.33}}$ \\ ${0.906_{\pm 0.027}}$ \\ ${33.72_{\pm 0.05}}$}  & \tabincell{c}{\cellcolor{gray!30}{${5.88_{\pm 0.01}}$} \\ \cellcolor{gray!30}{${0.920_{\pm 0.013}}$} \\  \cellcolor{gray!30}{${35.27_{\pm 0.02}}$}}\\
    \midrule
    Fire  & \tabincell{c}{MSE \\ SSIM ($ \uparrow $) \\ PSNR ($ \uparrow $)} & \tabincell{c}{${7.78_{\pm 0.24}}$ \\ ${0.891_{\pm 0.017}}$ \\ ${23.58_{\pm 0.03}}$} & \tabincell{c}{${4.41_{\pm 0.11}}$ \\ ${0.870_{\pm 0.019}}$ \\ ${27.55_{\pm 0.04}}$} & \tabincell{c}{${7.21_{\pm 0.20}}$ \\ ${0.901_{\pm 0.009}}$ \\ ${29.05_{\pm 0.03}}$} &  \tabincell{c}{${3.01_{\pm 0.16}}$ \\ ${0.903_{\pm 0.017}}$ \\ ${27.25_{\pm 0.04}}$} & \tabincell{c}{${2.19_{\pm 0.08}}$ \\ ${0.891_{\pm 0.007}}$ \\ ${29.03_{\pm 0.04}}$} & \tabincell{c}{${1.20_{\pm 0.04}}$ \\ ${0.917_{\pm 0.004}}$ \\ ${28.97_{\pm 0.03}}$}  & \tabincell{c}{\cellcolor{gray!30}${0.98_{\pm 0.02}}$ \\ \cellcolor{gray!30}${0.945_{\pm 0.009}}$ \\ \cellcolor{gray!30}${29.92_{\pm 0.02}}$}\\
    \midrule
    \tabincell{c}{SEVIR \\ (Ours-$L$)} & \tabincell{c}{MSE/10 ($ \downarrow $) \\ SSIM ($ \uparrow $) \\ PSNR ($ \uparrow $)} & \tabincell{c}{${10.54_{\pm 0.33}}$ \\ ${0.750_{\pm 0.018}}$ \\ ${27.55_{\pm 0.23}}$} & \tabincell{c}{${8.19_{\pm 0.34}}$ \\ ${0.870_{\pm 0.013}}$ \\ ${29.23_{\pm 0.37}}$} & \tabincell{c}{${6.73_{\pm 0.26}}$ \\ ${0.891_{\pm 0.008}}$ \\ ${27.25_{\pm 0.20}}$} & \tabincell{c}{${7.98_{\pm 0.32}}$ \\ ${0.870_{\pm 0.007}}$ \\ ${29.01_{\pm 0.16}}$} & \tabincell{c}{${5.82_{\pm 0.21}}$ \\ ${0.901_{\pm 0.005}}$ \\ ${28.97_{\pm 0.09}}$} & \tabincell{c}{${6.63_{\pm 0.26}}$ \\ ${0.902_{\pm 0.004}}$ \\ ${29.25_{\pm 0.20}}$}  & \tabincell{c}{\cellcolor{gray!30}${4.84_{\pm 0.17}}$ \\ \cellcolor{gray!30}${0.911_{\pm 0.005}}$ \\ \cellcolor{gray!30}${33.72_{\pm 0.07}}$}\\
    \bottomrule
  \end{tabular}
  \begin{tablenotes}
  \footnotesize
  \item[1]  \small {We take SSIM as the most representative index.}
   \end{tablenotes}
 \end{threeparttable}
 % \vspace{-1em}
\end{table*}

% \vspace{-0.6em}
\subsection{Performance on Real-world Datasets}\label{sec:4-1}

\begin{table}[h]\footnotesize
  \centering
  \renewcommand{\arraystretch}{0.80}
    \setlength{\tabcolsep}{8.0pt}
  \caption{The effect of the encoder and codebank size on SSIM. w/o denotes the remove of the codebank}
  % \vspace{-0.5em}
    \begin{tabular}{c|c|ccc}
    \toprule
    & \multirow{2}{*}{codebank size} & \multicolumn{3}{c}{Encoder size} \\
    & & Ours-$S$ & Ours-$B$ & Ours-$L$ \\
    \midrule
    \tabincell{c}{Reaction \\ -diffusion} & \tabincell{c}{ w/o \\ $128 \times 32$ \\ $256 \times 64$ \\ $512 \times 128$} & \tabincell{c}{ $0.765$\\ $0.877$ \\ $0.901$ \\ $0.867$} & \tabincell{c}{$0.765$\\ $0.784$ \\ $0.920$ \\ $0.892$} & \tabincell{c}{$0.741$\\ $0.817$ \\ $0.917$ \\ $0.817$} \\
    \midrule
    Fire  & \tabincell{c}{w/o \\ $128 \times 32$ \\ $256 \times 64$ \\ $512 \times 128$} & \tabincell{c}{$0.877$\\ $0.923$ \\ $0.927$ \\ $0.892$} & \tabincell{c}{$0.896$\\ $0.941$ \\ $0.945$ \\ $0.878$} & \tabincell{c}{$0.831$\\ $0.935$ \\ $0.939$ \\ $0.893$} \\
    \midrule
    \tabincell{c}{SEVIR} & \tabincell{c}{w/o \\ $128 \times 32$ \\ $256 \times 64$ \\ $512 \times 128$} & \tabincell{c}{$0.732$\\  $0.805$ \\ $0.867$ \\ $0.855$} & \tabincell{c}{$0.741$\\ $0.831$ \\ $0.901$ \\ $0.881$} & \tabincell{c}{$0.753$\\ $0.857$ \\ $0.911$ \\ $0.894$} \\

    % & \multirow{2}{*}{codebank size} & \multicolumn{3}{c}{Encoder size} \\
    % & & Ours-$S$ & Ours-$B$ & Ours-$L$ \\
    % \midrule
    % \tabincell{c}{Reaction \\ -diffusion} & \tabincell{c}{$128 \times 32$ \\ $256 \times 64$ \\ $512 \times 128$} & \tabincell{c}{$0.877$ \\ $0.901$ \\ $0.867$} & \tabincell{c}{$0.896$ \\ $0.920$ \\ $0.892$} & \tabincell{c}{$0.817$ \\ $0.917$ \\ $0.817$} \\
    % \midrule
    % Fire  & \tabincell{c}{$128 \times 32$ \\ $256 \times 64$ \\ $512 \times 128$} & \tabincell{c}{$0.923$ \\ $0.927$ \\ $0.892$} & \tabincell{c}{$0.941$ \\ $0.945$ \\ $0.878$} & \tabincell{c}{$0.935$ \\ $0.939$ \\ $0.893$} \\
    % \midrule
    % \tabincell{c}{SEVIR} & \tabincell{c}{$128 \times 32$ \\ $256 \times 64$ \\ $512 \times 128$} & \tabincell{c}{$0.805$ \\ $0.867$ \\ $0.855$} & \tabincell{c}{$0.831$ \\ $0.901$ \\ $0.881$} & \tabincell{c}{$0.857$ \\ $0.911$ \\ $0.894$} \\ 
    \bottomrule
  \end{tabular}
  \label{tab:ablation}%
\end{table}%

\textbf{Model comparison.} Tab.~\ref{tab:main} shows the results on the three physical datasets, where we adopt the base model on Reaction-diffusion and Fire datasets, and the large model on SEVIR. Primarily, our framework consistently surpasses the previous SOTAs by substantial performance margins across all physical datasets, \textit{e.g.}, performs 0.81$\sim$4.47 better than MathRel across three datasets on PSNR, and achieves 0.007$\sim$0.028 improvements on SSIM. For simplicity, we use SSIM as the major metric for further discussion. From Tab.~\ref{tab:main}, we also observe that while some physical methods may exhibit good performances on certain data, their performance may not be satisfactory in other scenarios. In particular, MathRel shows prominent performances on SEVIR and Reaction-diffusion, however, it encounters $\sim$0.1 deterioration on SSIM over the video understanding method PredRNN-v2. 

\begin{figure}[!h]
\centering  
\includegraphics[width=0.89\columnwidth]{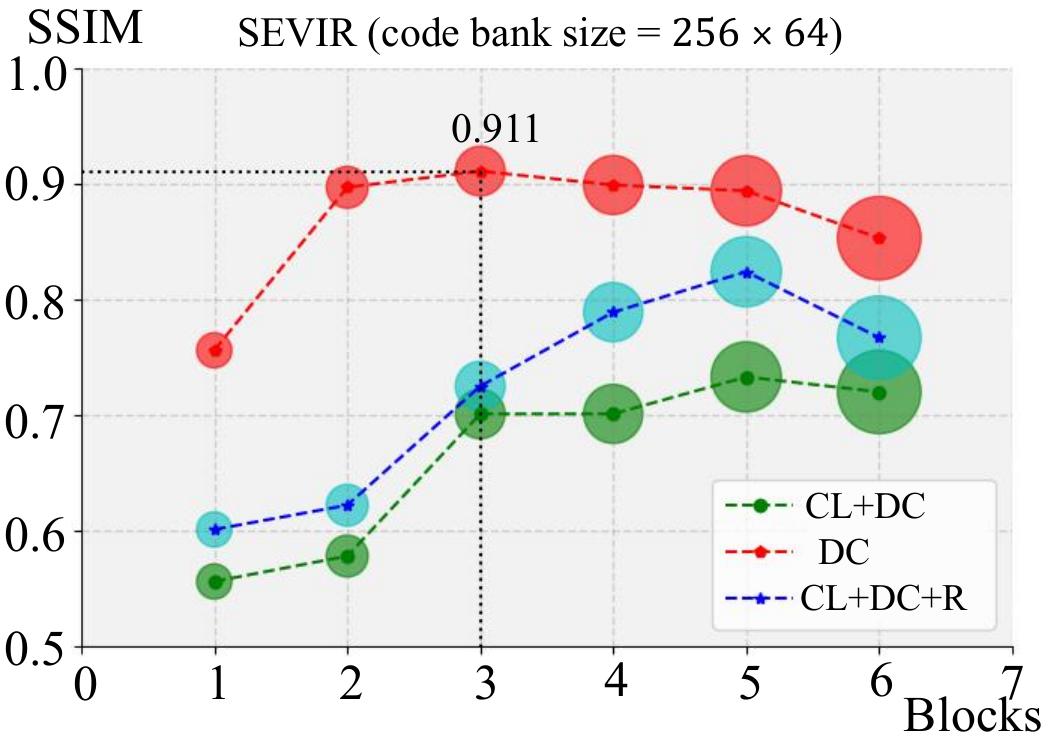}
% \vspace{-0.6em}
\caption{Effects of decoder forms and the decoder depth on SSIM. Datapoints are annotated with SSIM of the model and their size is proportional to areas.}
\label{ablation}
\end{figure}

\begin{figure}[!h]
\centering
\includegraphics[width=0.99\columnwidth]{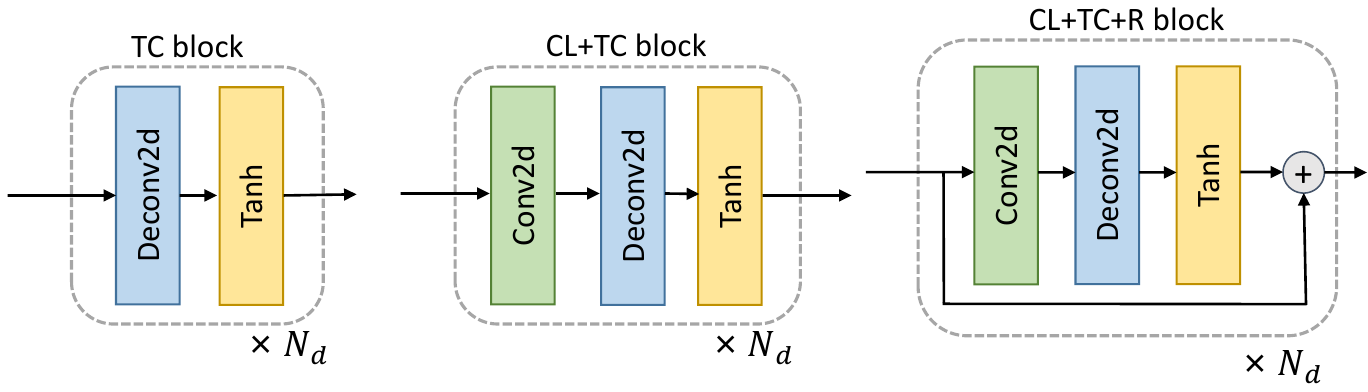}
% \vspace{-0.7em}
\caption{ The decoer forms of DC, CL+DC, CL+DC+R.
}
\label{fig:decoder form}
\end{figure}

\noindent \textbf{Scalability study.} We further evaluate the scalability of our framework, including different encoder sizes (Ours-$S/B/L$), decoder forms, and codebank sizes. We list our observations (Obs) as follows. \textbf{Obs.1}: Tab. \ref{tab:ablation} demonstrates that the size of the codebank should not be too large. We believe that an excessive dimension may increase the burden on the model, leading to the learning of noise. \textbf{Obs.2}: encoder sizes may need to vary based on the features of different datasets. As the spatial resolution of SEVIR dataset is 384$\times$384 (others are 128$\times$128) and this dataset has the largest data volume, it performs the best with our-$L$ design. In contrast, using smaller versions (\textit{i.e.}, our-$S/B$) performs better on the other two datasets. These facts validate that \textit{our encoder is highly scalable, enabling it to be customized for specific datasets, thereby significantly enhancing its real-world performance.} \textbf{Obs.3}: Compare with encoder size, improved performance can be attained by utilizing a moderate number of DC blocks, rather than an excessive amount of them. As shown in Fig. \ref{ablation}, under three blocks, CL+DC and CL+DC+R suffer $\sim 0.2$ deterioration on SSIM with only DC settings (about 0.9 \textit{vs.} 0.7). To strike a balance between computational resources and performance, we have opted for a solution that involves 3 DC blocks. See decoder detailed forms in Fig.~\ref{fig:decoder form}.

\begin{table*}[!t]\footnotesize
% \small
\setlength{\tabcolsep}{8.0pt}
  \caption{3-run results on MovingMNIST. The mean and standard deviation are reported.} 
  % \vspace{-0.5em}
  \label{tab:transfer}
  \centering
  \begin{tabular}{cc|ccccccccc}
    \toprule
    & Metrics & Ad-fusion & PhyDNet & PredRNN-V2 & MotionRNN & Earthformer & MathRel & Ours-$B$  & Ours-$L$ \\
    \midrule
    & PSNR & ${26.27_{\pm 0.31}}$ & ${29.99_{\pm 0.29}}$ & ${29.17_{\pm 0.26}}$ & ${30.19_{\pm 0.23}}$ & ${31.33_{\pm 0.21}}$ & ${37.17_{\pm 0.27}}$ & \cellcolor{gray!30}{${38.38_{\pm 0.17}}$} & \cellcolor{gray!30}{${37.46_{\pm 0.28}}$}  \\
    % & $L$ & -- & -- & -- & -- & -- & -- & \cellcolor{gray!30}{${37.46_{\pm 0.28}}$} 
    \midrule
    & SSIM & ${0.750_{\pm 0.04}}$ & ${0.869_{\pm 0.05}}$ & ${0.901_{\pm 0.06}}$ & ${0.891_{\pm 0.05}}$ & ${0.917_{\pm 0.07}}$ & ${0.937_{\pm 0.04}}$ & \cellcolor{gray!30}{${0.948_{\pm 0.03}}$}  & \cellcolor{gray!30}{${0.921_{\pm 0.01}}$} \\
    % & $L$ & -- & -- & -- & -- & -- & -- & \cellcolor{gray!30}{${0.921_{\pm 0.005}}$}  \\
    \bottomrule
  \end{tabular}
  % \vspace{-1em}
\end{table*}

\begin{figure}[!h]
\centering
\includegraphics[width=0.99\columnwidth]{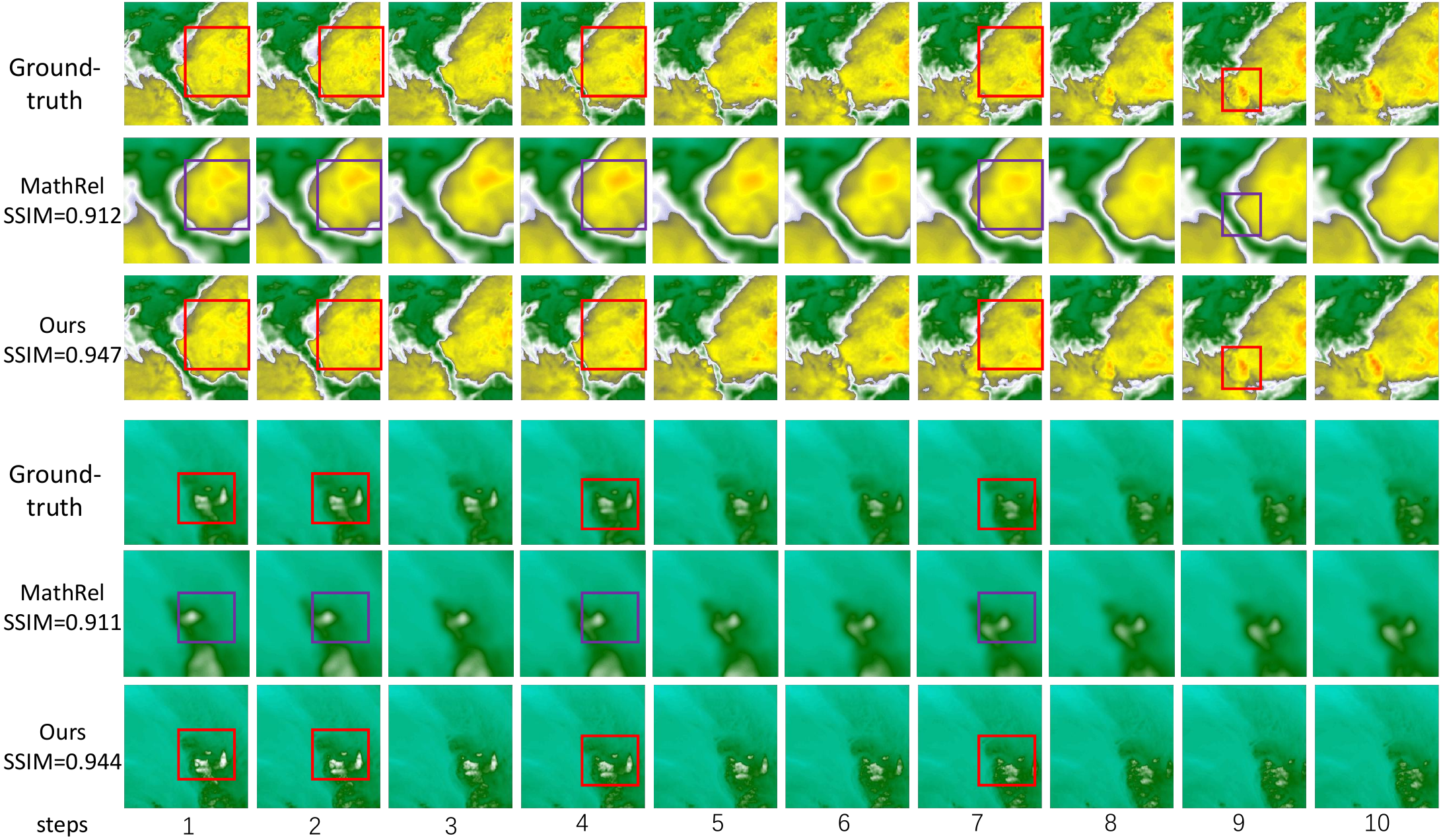}
% \vspace{-1.0em}
\caption{Visualization on SEVIR. }
\label{fig:sevir}
\end{figure}

\noindent \textbf{Visualization.} Fig.~\ref{fig:sevir} depicts the visualization results of our framework on real-world infrared satellite imagery dataset SEVIR, from which we can easily find that our model not only maintains the corresponding change characteristics globally, but also preserves local fidelity remarkably well.  This indicates that our model is capable of capturing both global and local spatiotemporal patterns in the data, making it a promising solution for a wide range of applications that require accurate and efficient modeling of complex spatiotemporal dynamics. 

\subsection{Results on MovingMNIST}\label{sec:4-2}
In this section, we provide more results to better examine our framework, we follow Earthformer \cite{gao2022earthformer} to use the public MovingMNIST dataset, and our target is to predict the future 10 frames for each sequence conditioned on the first 10 frames.

\begin{figure}[!t]
\centering
\includegraphics[width=0.99\columnwidth]{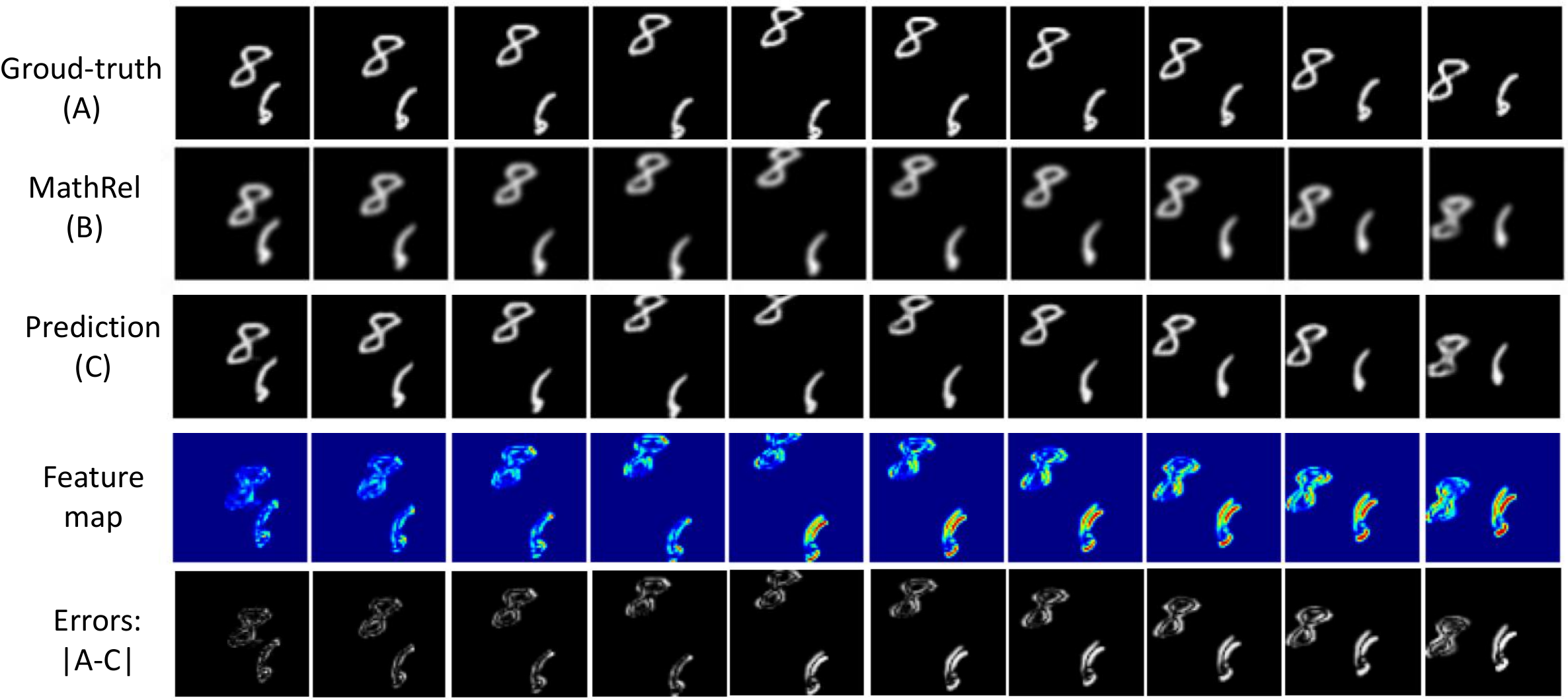}
\caption{Visualization on the synthetic dataset (MovingMNIST).}
\label{fig:moving}
% \vspace{-1em}
\end{figure}

\noindent \textbf{Model comparison.} The dynamics of the synthetic MovingMNIST dataset, in which the digits move independently with constant speed, are oversimplified when compared to the dynamics of complex physical systems. From Tab.~\ref{tab:transfer}, we can find that our frameworks (ours-$B/L$) are capable of learning long-range interactions among digits and accurately predicting their future motion trajectories. Furthermore, our model leverages global and local insights to more precisely predict the positions of the digits (achiving $\sim 1.21/0.011$ improvements on PSNR and SSIM, respectively). We further showcase the visualization examples of our framework with \textbf{SOTA} in Fig.~\ref{fig:moving}. Firstly, it excels in capturing structural information, outperforming existing models in local processing. Secondly, the feature map generated by the vector quantization prior bank effectively represents the original input information with both global and local fidelity.

\subsection{Ablation study} \textit{While our proposal is quite effective, we believe there still exists unapprehended questions.  We are eager to know:} ${{\cal Q}_1}:$ What is the role of local modules in enhancing local insights? ${{\cal Q}_2}:$ What roles do the general expert components play? ${{\cal Q}_3}:$ How the ${{\cal L}_{of}}$ influence the performance?

\noindent \textbf{Answer of ${{\cal Q}_1}$ and ${{\cal Q}_2}$.} As illustrated in Tab.~\ref{tab:local-level} and \ref{tab:general expert}, the model's performance experiences varying degrees of decline when the local pipeline and general expert components are removed. These results demonstrate the impressive performance of our model, which clearly benefits from the inclusion of both the local pipeline and general expert components. By leveraging these complementary sources of information, our model achieves superior accuracy and robustness across a range of tasks and domains, making it a highly promising solution for real-world applications.

\begin{table}[h]\footnotesize
  \centering
  \renewcommand{\arraystretch}{0.80}
    \setlength{\tabcolsep}{8.0pt}
  \caption{Ablations of local-level pipeline.}
  % \vspace{-0.5em}
    \begin{tabular}{c|cc}
    \toprule
          & \multicolumn{2}{c}{Top (Local) level pipeline (SSIM)} \\
    \midrule
    Reaction-diffusion ($B$) & ${0.920_{\pm 0.013}}$ (\textcolor{blue}{$\checkmark$}) & ${0.911_{\pm 0.021}}$ (\textcolor{blue}{$\times$}) \\
    Fire ($B$) & ${0.945_{\pm 0.009}}$ (\textcolor{blue}{$\checkmark$}) & ${0.935_{\pm 0.011}}$(\textcolor{blue}{$\times$}) \\
    SEVIR ($L$) & ${0.911_{\pm 0.005}}$ (\textcolor{blue}{$\checkmark$}) & ${0.897_{\pm 0.007}}$ (\textcolor{blue}{$\times$}) \\
    MovingMNIST ($B$) & ${0.921_{\pm 0.005}}$ (\textcolor{blue}{$\checkmark$}) & ${0.918_{\pm 0.006}}$ (\textcolor{blue}{$\times$}) \\
    \bottomrule
    \end{tabular}%
  \label{tab:local-level}
\end{table}%

\begin{table}[h]\footnotesize
  \centering
  \renewcommand{\arraystretch}{0.80}
    \setlength{\tabcolsep}{8.0pt}
  \caption{Implementation details on four datasets.}
  % \vspace{-0.5em}
\begin{tabular}{c|cc}
    \toprule
          & \multicolumn{2}{c}{Ablations of general expert components.} \\
    \midrule
    Reaction-diffusion ($B$) & ${0.920_{\pm 0.013}}$ (\textcolor{blue}{$\checkmark$}) & ${0.897_{\pm 0.031}}$ (\textcolor{blue}{$\times$}) \\
    Fire ($B$) & ${0.945_{\pm 0.009}}$ (\textcolor{blue}{$\checkmark$}) & ${0.917_{\pm 0.021}}$ (\textcolor{blue}{$\times$}) \\
    SEVIR ($L$) & ${0.911_{\pm 0.005}}$ (\textcolor{blue}{$\checkmark$}) & ${0.893_{\pm 0.011}}$ (\textcolor{blue}{$\times$}) \\
    MovingMNIST ($B$) & ${0.921_{\pm 0.005}}$ (\textcolor{blue}{$\checkmark$}) & ${0.903_{\pm 0.012}}$ (\textcolor{blue}{$\times$}) \\
    \bottomrule
    \end{tabular}%
 \label{tab:general expert}
\end{table}%

\noindent \textbf{Answer of ${{\cal Q}_3}$.} As illustrated in Tab \ref{tab:optical}, we choose SOTAs compare with our model (without (w/o) ${{\cal L}_{of}}$) to verify the influence of optical flow estimation module. The results demonstrate a clear decline in both SSIM and PSNR values when the optical flow estimation module is removed. The reduction in these key metrics signifies a degradation in the model's ability to maintain structural integrity and signal quality. These findings reaffirm our decision to incorporate this module.

\begin{table}[h]\scriptsize
% \small
\setlength{\tabcolsep}{1.3pt}
  \caption{3-run results on SEVIR. The mean and standard deviation are reported.} 
  % \vspace{-0.5em}
  \label{tab:optical}
  \centering
  \begin{tabular}{cc|ccccccccc}
    \toprule
    & Metrics  & PredRNN-V2 & Earthformer & MathRel & Ours-$L$  & Ours-$L$ w/o ${{\cal L}_{of}}$ \\
    \midrule
    & PSNR & ${29.01_{\pm 0.16}}$ & ${28.97_{\pm 0.09}}$ & ${29.25_{\pm 0.20}}$ & ${33.72_{\pm 0.07}}$ & ${30.14_{\pm 0.09}}$  \\
    % & $L$ & -- & -- & -- & -- & -- & -- & \cellcolor{gray!30}{${37.46_{\pm 0.28}}$} 
    \midrule
    & SSIM & ${0.891_{\pm 0.08}}$ & ${0.901_{\pm 0.005}}$ & ${0.902_{\pm 0.04}}$ & ${0.911_{\pm 0.005}}$ & ${0.875_{\pm 0.011}}$  \\
    % & $L$ & -- & -- & -- & -- & -- & -- & \cellcolor{gray!30}{${0.921_{\pm 0.005}}$}  \\
    \bottomrule
  \end{tabular}
  % \vspace{-1em}
\end{table}

% \vspace{-0.5em}
\subsection{Compare with video understanding frameworks}
We further explored various video understanding benchmarks, including MIM \cite{wang2019memory} (CVPR 2019), CrevNet \cite{yu2020efficient} (ICLR 2020), and SimVP \cite{gao2022simvp} (CVPR 2022). Here we choose SEVIR and MovingMNIST as benchmarks. As depicted in Tab.~\ref{tab:add}, on the MovingMNIST dataset, SimVP, introduced at CVPR 2022, yielded slightly better results than MIM and CrevNet, with a PSNR of 38.02 and an SSIM score of 0.939. Notably, our method, denoted as Ours-$B$, surpassed all the above, reaching a PSNR of 38.01 and an SSIM score of 0.945. On SEVIR, our method, designated as Ours-$L$, outperformed the mainstream models by achieving a significant PSNR of 33.72 and an SSIM score of 0.911. These results not only underscore the advanced performance of our proposed framework but also highlight its superiority over existing mainstream methodologies in video understanding.

\begin{table}[htbp]\footnotesize
  \centering
    \renewcommand{\arraystretch}{0.80}
  \setlength{\tabcolsep}{8.5pt}
  \caption{Comparison among different and additional video understanding baselines across MovingMNIST and SEVIR datasets. To simplify comprehension, we run three times experiments and report the results of SSIM and PSNR.}
  % \vspace{-0.5em}
    \begin{tabular}{ccccc}
    \toprule
    Dataset & Method & Conference & PSNR  & SSIM \\
    \midrule
    \multirow{4}[2]{*}{MovingMNIST} & MIM   & CVPR 2019 & 35.98 & 0.909 \\
          & CrevNet & ICLR 2020 & 37.69 & 0.934 \\
          & SimVP & CVPR 2022 & 38.02 & 0.939 \\
          & Ours-$B$ & --    & 38.01 & 0.945 \\
    \midrule
    \multirow{4}[2]{*}{SEVIR} & MIM   & CVPR 2019 & 29.99 & 0.854 \\
          & CrevNet & ICLR 2020 & 31.45 & 0.895 \\
          & SimVP & CVPR 2022 & 31.88 & 0.897 \\
          & Ours-$L$ & --    & 33.72 & 0.911 \\
    \bottomrule
    \end{tabular}%
  \label{tab:add}%
\end{table}%

\subsection{Additional experiments on SEVIR with Fourier-based Neural Network}

In this part, we turn to quantificationally analyze our model generalizability compare with specialized models \cite{wu2023pastnet, guibas2021adaptive}, which  specifically designed for addressing continuous and PDE problems. These models include Fourier Neural Operator (FNO) \cite{li2020fourier}, Adaptive Fourier Neural Operator (AFNO) \cite{guibas2021adaptive}, and TF-Net \cite{wang2020towards}. All three models leverage the properties of the Fourier transform. The Fourier transform is a technique that converts functions from the time domain (or spatial domain) to the frequency domain, enabling these models to process problems in the frequency domain and utilize certain mathematical properties of the Fourier transform. We showcase the results in Tab \ref{tab:fourier}.

\begin{table}[h]\footnotesize
  \centering
  \caption{Comparison among different and additional Fourier-based network baselines across SEVIR. To simplify comprehension, we run three times experiments and report the results of SSIM and PSNR.}
  % \vspace{-0.5em}
    \begin{tabular}{ccccc}
    \toprule
    Dataset & Method & MSE & SSIM  & PSNR \\
    \midrule
    \multirow{4}[2]{*}{SEVIR} & AFNO    & ${15.87_{\pm 0.89}}$ & ${0.521_{\pm 0.009}}$  & ${23.43_{\pm 0.18}}$  \\
          & TF-Net & ${8.47_{\pm 0.33}}$  & ${0.774_{\pm 0.011}}$  & ${30.12_{\pm 0.23}}$  \\
          & FNO & ${12.21_{\pm 0.75}}$  & ${0.583_{\pm 0.012}}$  & ${26.43_{\pm 0.13}}$  \\
          & PhysicNet ($L$) & ${4.84_{\pm 0.17}}$  & ${0.911_{\pm 0.005}}$  & ${33.72_{\pm 0.07}}$  \\
    \bottomrule
    \end{tabular}%
  \label{tab:fourier}%
\end{table}%

Based on the comparison presented in Table \ref{tab:fourier}, we can observe that our proposed model, PhysicNet ($L$), consistently outperforms other networks across all metrics on the SEVIR dataset. The Mean Square Error (MSE) achieved by PhysicNet is significantly lower, indicating a superior predictive accuracy. Additionally, PhysicNet's Structural Similarity Index Measure (SSIM) is the highest among the models, indicating that it captures the structural detail in the prediction more effectively. Similarly, PhysicNet also delivers the highest Peak Signal-to-Noise Ratio (PSNR), reflecting its superior ability to maintain signal quality amidst potential noise.

The results demonstrate the exceptional generalization capability of PhysicNet in comparison to other specialized models specifically designed for continuous and PDE problems. While Fourier-based methods, like AFNO, FNO, and TF-Net, leverage the mathematical properties of Fourier transforms to work in the frequency domain, PhysicNet appears to be more efficient and accurate in capturing and predicting complex patterns, possibly due to its design and learning algorithm. In summary, PhysicNet represents a substantial step forward in addressing complex continuous and PDE problems, offering a promising approach for future research and practical applications.

\subsection{Visual Comparison of Model Errors}

In this section, we further delve into an analysis through visualization to assess our model's capability in capturing fine-grained details. We showcase the visual errors of our model and compare them with those of current mainstream frameworks. The weather dataset, SEVIR, is utilized as a benchmark given its high-resolution pixel data and rich dimensional information.

\begin{figure}[h]
\centering
\includegraphics[width=0.99\columnwidth]{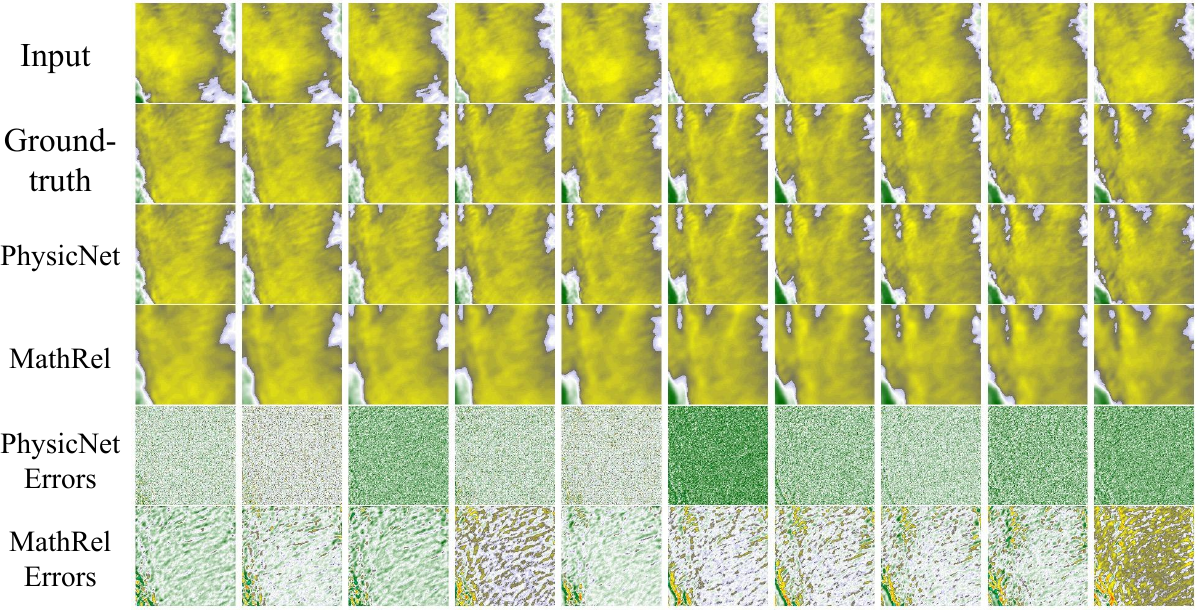}
\caption{Visualization on real-world infrared satellite imagery dataset (SEVIR).}
\label{fig:sevir_app1}
\end{figure}

\begin{figure}[h]
\centering
\includegraphics[width=0.99\columnwidth]{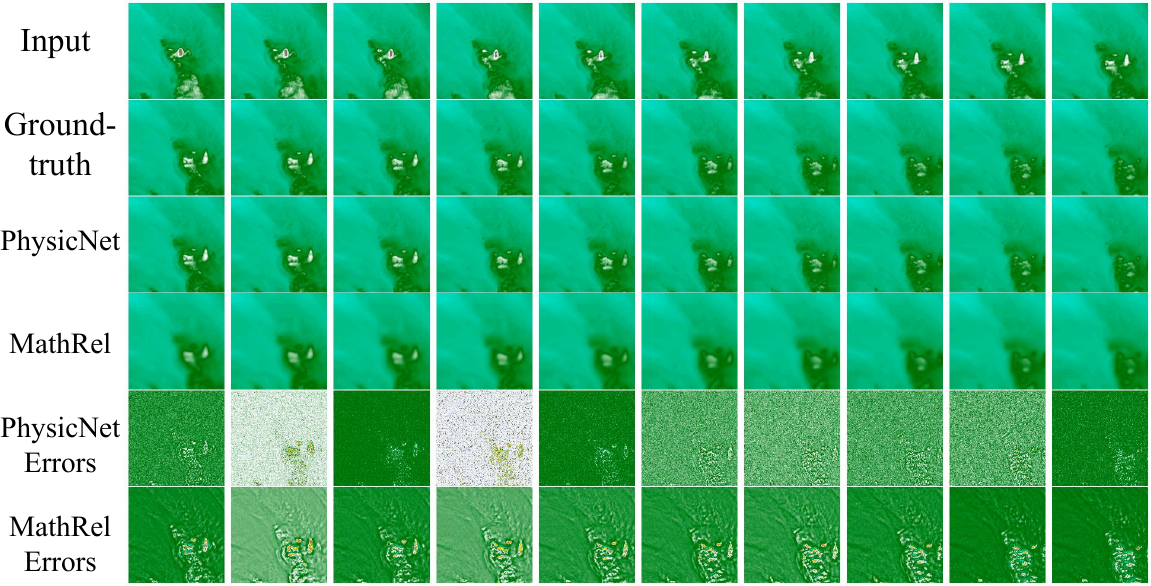}
\caption{Visualization on real-world infrared satellite imagery dataset (SEVIR).}
\label{fig:sevir_app2}
\end{figure}

From the visualization results, it is readily evident that our model exhibits a smaller prediction error compared to MathRel. Furthermore, our approach handles fine details more adeptly. As observed in Fig \ref{fig:sevir_app1} and \ref{fig:sevir_app2}, although MathRel already achieves commendable prediction outcomes, it still falls short in capturing local details, especially at the outline information. In contrast, PhysicNet adeptly addresses this limitation, which corroborates our hypothesis of joint global and local modeling.

\section{Complexity analysis}

To better validate the efficiency of the PhysicNet algorithm, we include efficiency comparisons in the manuscript, which aid in a deeper understanding of our algorithm's reliability and effectiveness. As shown in Tab \ref{tab:complex}, we observe significant variations in memory usage, computational complexity (FLOPs), the number of parameters (in millions), and the training time per epoch. The experimental analysis demonstrates that our PhysicNet model is an effective algorithm among the compared models, achieving a good balance between resource usage and computational performance. It utilizes less memory and has lower computational complexity than the most demanding model, MIM, while maintaining a competitive training time. This efficiency suggests that PhysicNet is an advantageous choice for tasks requiring a judicious balance of speed and computational resources.

\begin{table}[h] \footnotesize
\centering
  \setlength{\tabcolsep}{2pt}
\caption{Comparison of different models in terms of memory usage, computational complexity (FLOPs), number of parameters (Params), and training time per epoch.}
\label{tab:complex}
\begin{tabular}{ccccc}
\toprule
\textbf{Model} & \textbf{Memory (MB)} & \textbf{FLOPs (G)} & \textbf{Params (M)} & \textbf{Training time} \\
\midrule
FNO & 8.41 & 12.31 & 7.27 & 32s / epoch \\
MIM & 2331.34 & 179.21 & 38.44 & 154s / epoch \\
PredRNN-V2 & 1721.22 & 117.33 & 23.92 & 126s / epoch \\
SimVP & 421.17 & 17.24 & 46.82 & 65s / epoch \\
PhysicNet & 378.87 & 15.86 & 35.41 & 58s / epoch \\
\bottomrule
\end{tabular}
\end{table}

\section{Conclusion \& Future work}\label{sec:5}
In conclusion, our model proposes a vector quantization-based discrete learning technique, combined with an expert module that captures the underlying patterns of motion in order to model physical processes. By leveraging these key components, our model achieves superior performance on a variety of challenging tasks, highlighting the effectiveness and robustness of our approach. We believe that our model has the potential to significantly advance the field of machine learning and enable new breakthroughs in our understanding of complex physical systems. We emphasize that the work presented in this paper is primarily foundational in ST context.

Although our work has shown promising results on real-world physical datasets, there are still some issues related to interpretability that need to be further addressed.  In particular, we plan to explore the use of manifold learning techniques in our model and conduct more experiments on physical data in other fields, which could help to inject greater interpretability into our results.  By leveraging these methods and datasets, we hope to better understand the underlying factors that drive the performance of our model and gain new insights into the physical systems.  

\section{Acknowledgment}

This paper is partially supported by the National Natural Science Foundation of China (No.62072427, No.12227901), the Project of Stable Support for Youth Team in Basic Research Field, CAS (No.YSBR-005), Academic Leaders Cultivation Program, USTC.

% or


\bibliographystyle{IEEEtran}
{
\bibliography{reference}

\begin{thebibliography}{100}
\providecommand{\url}[1]{#1}
\csname url@samestyle\endcsname
\providecommand{\newblock}{\relax}
\providecommand{\bibinfo}[2]{#2}
\providecommand{\BIBentrySTDinterwordspacing}{\spaceskip=0pt\relax}
\providecommand{\BIBentryALTinterwordstretchfactor}{4}
\providecommand{\BIBentryALTinterwordspacing}{\spaceskip=\fontdimen2\font plus
\BIBentryALTinterwordstretchfactor\fontdimen3\font minus
  \fontdimen4\font\relax}
\providecommand{\BIBforeignlanguage}[2]{{%
\expandafter\ifx\csname l@#1\endcsname\relax
\typeout{** WARNING: IEEEtran.bst: No hyphenation pattern has been}%
\typeout{** loaded for the language `#1'. Using the pattern for}%
\typeout{** the default language instead.}%
\else
\language=\csname l@#1\endcsname
\fi
#2}}
\providecommand{\BIBdecl}{\relax}
\BIBdecl

\bibitem{benacerraf1973mathematical}
P.~Benacerraf, ``Mathematical truth,'' \emph{The Journal of Philosophy},
  vol.~70, no.~19, pp. 661--679, 1973.

\bibitem{newell1980physical}
A.~Newell, ``Physical symbol systems,'' \emph{Cognitive science}, vol.~4,
  no.~2, pp. 135--183, 1980.

\bibitem{pryor2009multiphysics}
R.~W. Pryor, \emph{Multiphysics modeling using COMSOL{\textregistered}: a first
  principles approach}.\hskip 1em plus 0.5em minus 0.4em\relax Jones \&
  Bartlett Publishers, 2009.

\bibitem{roberts2022principles}
D.~A. Roberts, S.~Yaida, and B.~Hanin, \emph{The principles of deep learning
  theory}.\hskip 1em plus 0.5em minus 0.4em\relax Cambridge University Press
  Cambridge, MA, USA, 2022.

\bibitem{burkle2021deep}
M.~B{\"u}rkle, U.~Perera, F.~Gimbert, H.~Nakamura, M.~Kawata, and Y.~Asai,
  ``Deep-learning approach to first-principles transport simulations,''
  \emph{Physical Review Letters}, vol. 126, no.~17, p. 177701, 2021.

\bibitem{yu2020learning}
Y.~Yu, K.~H.~R. Chan, C.~You, C.~Song, and Y.~Ma, ``Learning diverse and
  discriminative representations via the principle of maximal coding rate
  reduction,'' \emph{Advances in Neural Information Processing Systems},
  vol.~33, pp. 9422--9434, 2020.

\bibitem{chan2022redunet}
K.~H.~R. Chan, Y.~Yu, C.~You, H.~Qi, J.~Wright, and Y.~Ma, ``Redunet: A
  white-box deep network from the principle of maximizing rate reduction,''
  \emph{The Journal of Machine Learning Research}, vol.~23, no.~1, pp.
  4907--5009, 2022.

\bibitem{anderson1972more}
P.~W. Anderson, ``More is different: broken symmetry and the nature of the
  hierarchical structure of science.'' \emph{Science}, vol. 177, no. 4047, pp.
  393--396, 1972.

\bibitem{pierson1993corpore}
S.~Pierson, ``Corpore cadente...: Historians discuss newton’s second law,''
  \emph{Perspectives on Science}, vol.~1, no.~4, pp. 627--658, 1993.

\bibitem{sharan1996mathematical}
M.~Sharan, A.~K. Yadav, M.~Singh, P.~Agarwal, and S.~Nigam, ``A mathematical
  model for the dispersion of air pollutants in low wind conditions,''
  \emph{Atmospheric Environment}, vol.~30, no.~8, pp. 1209--1220, 1996.

\bibitem{egan1972numerical}
B.~A. Egan and J.~R. Mahoney, ``Numerical modeling of advection and diffusion
  of urban area source pollutants,'' \emph{Journal of Applied Meteorology and
  Climatology}, vol.~11, no.~2, pp. 312--322, 1972.

\bibitem{shan2006kinetic}
X.~Shan, X.-F. Yuan, and H.~Chen, ``Kinetic theory representation of
  hydrodynamics: a way beyond the navier--stokes equation,'' \emph{Journal of
  Fluid Mechanics}, vol. 550, pp. 413--441, 2006.

\bibitem{mccracken2018artificial}
M.~F. McCracken, ``Artificial neural networks in fluid dynamics: A novel
  approach to the navier-stokes equations,'' in \emph{Proceedings of the
  Practice and Experience on Advanced Research Computing}, 2018, pp. 1--4.

\bibitem{arrowsmith1992dynamical}
D.~Arrowsmith and C.~M. Place, \emph{Dynamical systems: differential equations,
  maps, and chaotic behaviour}.\hskip 1em plus 0.5em minus 0.4em\relax CRC
  Press, 1992, vol.~5.

\bibitem{hirsch1984dynamical}
M.~W. Hirsch, ``The dynamical systems approach to differential equations,''
  \emph{Bulletin of the American mathematical society}, vol.~11, no.~1, pp.
  1--64, 1984.

\bibitem{hirsch2012differential}
M.~W. Hirsch, S.~Smale, and R.~L. Devaney, \emph{Differential equations,
  dynamical systems, and an introduction to chaos}.\hskip 1em plus 0.5em minus
  0.4em\relax Academic press, 2012.

\bibitem{sideris2013ordinary}
T.~C. Sideris, \emph{Ordinary differential equations and dynamical
  systems}.\hskip 1em plus 0.5em minus 0.4em\relax Springer, 2013, vol.~2.

\bibitem{verhulst2006nonlinear}
F.~Verhulst, \emph{Nonlinear differential equations and dynamical
  systems}.\hskip 1em plus 0.5em minus 0.4em\relax Springer Science \& Business
  Media, 2006.

\bibitem{wiggins2003introduction}
S.~Wiggins, S.~Wiggins, and M.~Golubitsky, \emph{Introduction to applied
  nonlinear dynamical systems and chaos}.\hskip 1em plus 0.5em minus
  0.4em\relax Springer, 2003, vol.~2, no.~3.

\bibitem{hale2012dynamics}
J.~K. Hale and H.~Ko{\c{c}}ak, \emph{Dynamics and bifurcations}.\hskip 1em plus
  0.5em minus 0.4em\relax Springer Science \& Business Media, 2012, vol.~3.

\bibitem{humar2012dynamics}
J.~Humar, \emph{Dynamics of structures}.\hskip 1em plus 0.5em minus 0.4em\relax
  CRC press, 2012.

\bibitem{wang2020physics}
L.~Wang, Q.~Zhou, and S.~Jin, ``Physics-guided deep learning for power system
  state estimation,'' \emph{Journal of Modern Power Systems and Clean Energy},
  vol.~8, no.~4, pp. 607--615, 2020.

\bibitem{harish2016review}
V.~Harish and A.~Kumar, ``A review on modeling and simulation of building
  energy systems,'' \emph{Renewable and sustainable energy reviews}, vol.~56,
  pp. 1272--1292, 2016.

\bibitem{moin1998direct}
P.~Moin and K.~Mahesh, ``Direct numerical simulation: a tool in turbulence
  research,'' \emph{Annual review of fluid mechanics}, vol.~30, no.~1, pp.
  539--578, 1998.

\bibitem{rogallo1984numerical}
R.~S. Rogallo and P.~Moin, ``Numerical simulation of turbulent flows,''
  \emph{Annual review of fluid mechanics}, vol.~16, no.~1, pp. 99--137, 1984.

\bibitem{orszag1972numerical}
S.~A. Orszag and G.~Patterson~Jr, ``Numerical simulation of three-dimensional
  homogeneous isotropic turbulence,'' \emph{Physical review letters}, vol.~28,
  no.~2, p.~76, 1972.

\bibitem{sanyal1999numerical}
J.~Sanyal, S.~V{\'a}squez, S.~Roy, and M.~Dudukovic, ``Numerical simulation of
  gas--liquid dynamics in cylindrical bubble column reactors,'' \emph{Chemical
  Engineering Science}, vol.~54, no.~21, pp. 5071--5083, 1999.

\bibitem{van2008numerical}
M.~A. van~der Hoef, M.~van Sint~Annaland, N.~Deen, and J.~Kuipers, ``Numerical
  simulation of dense gas-solid fluidized beds: a multiscale modeling
  strategy,'' \emph{Annu. Rev. Fluid Mech.}, vol.~40, pp. 47--70, 2008.

\bibitem{yang2011experimental}
P.~Yang, X.~Tan, and W.~Xin, ``Experimental study and numerical simulation for
  a storehouse fire accident,'' \emph{Building and Environment}, vol.~46,
  no.~7, pp. 1445--1459, 2011.

\bibitem{ma2003numerical}
T.~Ma and J.~Quintiere, ``Numerical simulation of axi-symmetric fire plumes:
  accuracy and limitations,'' \emph{Fire Safety Journal}, vol.~38, no.~5, pp.
  467--492, 2003.

\bibitem{bakarji2022discovering}
J.~Bakarji, K.~Champion, J.~N. Kutz, and S.~L. Brunton, ``Discovering governing
  equations from partial measurements with deep delay autoencoders,''
  \emph{arXiv preprint arXiv:2201.05136}, 2022.

\bibitem{he2021deep}
X.~He, Q.~He, and J.-S. Chen, ``Deep autoencoders for physics-constrained
  data-driven nonlinear materials modeling,'' \emph{Computer Methods in Applied
  Mechanics and Engineering}, vol. 385, p. 114034, 2021.

\bibitem{cannon2003dynamics}
R.~H. Cannon, \emph{Dynamics of physical systems}.\hskip 1em plus 0.5em minus
  0.4em\relax Courier Corporation, 2003.

\bibitem{bezruchko2010extracting}
B.~P. Bezruchko and D.~A. Smirnov, \emph{Extracting knowledge from time series:
  An introduction to nonlinear empirical modeling}.\hskip 1em plus 0.5em minus
  0.4em\relax Springer Science \& Business Media, 2010.

\bibitem{chen2022automated}
B.~Chen, K.~Huang, S.~Raghupathi, I.~Chandratreya, Q.~Du, and H.~Lipson,
  ``Automated discovery of fundamental variables hidden in experimental data,''
  \emph{Nature Computational Science}, vol.~2, no.~7, pp. 433--442, 2022.

\bibitem{godunov1959finite}
S.~K. Godunov and I.~Bohachevsky, ``Finite difference method for numerical
  computation of discontinuous solutions of the equations of fluid dynamics,''
  \emph{Matemati{\v{c}}eskij sbornik}, vol.~47, no.~3, pp. 271--306, 1959.

\bibitem{bar1999fitting}
M.~B{\"a}r, R.~Hegger, and H.~Kantz, ``Fitting partial differential equations
  to space-time dynamics,'' \emph{Physical Review E}, vol.~59, no.~1, p. 337,
  1999.

\bibitem{yanchuk2017spatio}
S.~Yanchuk and G.~Giacomelli, ``Spatio-temporal phenomena in complex systems
  with time delays,'' \emph{Journal of Physics A: Mathematical and
  Theoretical}, vol.~50, no.~10, p. 103001, 2017.

\bibitem{baker1997mathematical}
C.~Baker, G.~Bocharov, and C.~Paul, ``Mathematical modelling of the
  interleukin--2 t--cell system: a comparative study of approaches based on
  ordinary and delay differential equation,'' \emph{Computational and
  Mathematical Methods in Medicine}, vol.~1, no.~2, pp. 117--128, 1997.

\bibitem{graves2012long}
A.~Graves and A.~Graves, ``Long short-term memory,'' \emph{Supervised sequence
  labelling with recurrent neural networks}, pp. 37--45, 2012.

\bibitem{mnih2013playing}
V.~Mnih, K.~Kavukcuoglu, D.~Silver, A.~Graves, I.~Antonoglou, D.~Wierstra, and
  M.~Riedmiller, ``Playing atari with deep reinforcement learning,''
  \emph{arXiv preprint arXiv:1312.5602}, 2013.

\bibitem{krizhevsky2017imagenet}
A.~Krizhevsky, I.~Sutskever, and G.~E. Hinton, ``Imagenet classification with
  deep convolutional neural networks,'' \emph{Communications of the ACM},
  vol.~60, no.~6, pp. 84--90, 2017.

\bibitem{creswell2018generative}
A.~Creswell, T.~White, V.~Dumoulin, K.~Arulkumaran, B.~Sengupta, and A.~A.
  Bharath, ``Generative adversarial networks: An overview,'' \emph{IEEE signal
  processing magazine}, vol.~35, no.~1, pp. 53--65, 2018.

\bibitem{he2016deep}
K.~He, X.~Zhang, S.~Ren, and J.~Sun, ``Deep residual learning for image
  recognition,'' in \emph{Proceedings of the IEEE conference on computer vision
  and pattern recognition}, 2016, pp. 770--778.

\bibitem{shi2015convolutional}
X.~Shi, Z.~Chen, H.~Wang, D.-Y. Yeung, W.-K. Wong, and W.-c. Woo,
  ``Convolutional lstm network: A machine learning approach for precipitation
  nowcasting,'' \emph{Advances in neural information processing systems},
  vol.~28, 2015.

\bibitem{dedeep}
E.~de~Bezenac, A.~Pajot, and P.~Gallinari, ``Deep learning for physical
  processes: Incorporating prior scientific knowledge,'' in \emph{International
  Conference on Learning Representations}.

\bibitem{guen2020disentangling}
V.~L. Guen and N.~Thome, ``Disentangling physical dynamics from unknown factors
  for unsupervised video prediction,'' in \emph{Proceedings of the IEEE/CVF
  Conference on Computer Vision and Pattern Recognition}, 2020, pp.
  11\,474--11\,484.

\bibitem{seo2020physics}
S.~Seo, C.~Meng, and Y.~Liu, ``Physics-aware difference graph networks for
  sparsely-observed dynamics,'' in \emph{International conference on learning
  representations}, 2020.

\bibitem{raissi2019physics}
M.~Raissi, P.~Perdikaris, and G.~E. Karniadakis, ``Physics-informed neural
  networks: A deep learning framework for solving forward and inverse problems
  involving nonlinear partial differential equations,'' \emph{Journal of
  Computational physics}, vol. 378, pp. 686--707, 2019.

\bibitem{karniadakis2021physics}
G.~E. Karniadakis, I.~G. Kevrekidis, L.~Lu, P.~Perdikaris, S.~Wang, and
  L.~Yang, ``Physics-informed machine learning,'' \emph{Nature Reviews
  Physics}, vol.~3, no.~6, pp. 422--440, 2021.

\bibitem{gao2022earthformer}
Z.~Gao, X.~Shi, H.~Wang, Y.~Zhu, Y.~B. Wang, M.~Li, and D.-Y. Yeung,
  ``Earthformer: Exploring space-time transformers for earth system
  forecasting,'' \emph{Advances in Neural Information Processing Systems},
  vol.~35, pp. 25\,390--25\,403, 2022.

\bibitem{wang2022predrnn}
Y.~Wang, H.~Wu, J.~Zhang, Z.~Gao, J.~Wang, S.~Y. Philip, and M.~Long,
  ``Predrnn: A recurrent neural network for spatiotemporal predictive
  learning,'' \emph{IEEE Transactions on Pattern Analysis and Machine
  Intelligence}, vol.~45, no.~2, pp. 2208--2225, 2022.

\bibitem{DBLP:conf/iclr/PfaffFSB21}
\BIBentryALTinterwordspacing
T.~Pfaff, M.~Fortunato, A.~Sanchez{-}Gonzalez, and P.~W. Battaglia, ``Learning
  mesh-based simulation with graph networks,'' in \emph{9th International
  Conference on Learning Representations, {ICLR} 2021, Virtual Event, Austria,
  May 3-7, 2021}.\hskip 1em plus 0.5em minus 0.4em\relax OpenReview.net, 2021.
  [Online]. Available: \url{https://openreview.net/forum?id=roNqYL0\_XP}
\BIBentrySTDinterwordspacing

\bibitem{van2017neural}
A.~Van Den~Oord, O.~Vinyals \emph{et~al.}, ``Neural discrete representation
  learning,'' \emph{Advances in neural information processing systems},
  vol.~30, 2017.

\bibitem{fortuin2018som}
V.~Fortuin, M.~H{\"u}ser, F.~Locatello, H.~Strathmann, and G.~R{\"a}tsch,
  ``Som-vae: Interpretable discrete representation learning on time series,''
  \emph{arXiv preprint arXiv:1806.02199}, 2018.

\bibitem{caron2021emerging}
M.~Caron, H.~Touvron, I.~Misra, H.~J{\'e}gou, J.~Mairal, P.~Bojanowski, and
  A.~Joulin, ``Emerging properties in self-supervised vision transformers,'' in
  \emph{Proceedings of the IEEE/CVF international conference on computer
  vision}, 2021, pp. 9650--9660.

\bibitem{stewart2017label}
R.~Stewart and S.~Ermon, ``Label-free supervision of neural networks with
  physics and domain knowledge,'' in \emph{Proceedings of the AAAI Conference
  on Artificial Intelligence}, vol.~31, no.~1, 2017.

\bibitem{watters2017visual}
N.~Watters, D.~Zoran, T.~Weber, P.~Battaglia, R.~Pascanu, and A.~Tacchetti,
  ``Visual interaction networks: Learning a physics simulator from video,''
  \emph{Advances in neural information processing systems}, vol.~30, 2017.

\bibitem{kashinath2021physics}
K.~Kashinath, M.~Mustafa, A.~Albert, J.~Wu, C.~Jiang, S.~Esmaeilzadeh,
  K.~Azizzadenesheli, R.~Wang, A.~Chattopadhyay, A.~Singh \emph{et~al.},
  ``Physics-informed machine learning: case studies for weather and climate
  modelling,'' \emph{Philosophical Transactions of the Royal Society A}, vol.
  379, no. 2194, p. 20200093, 2021.

\bibitem{wang2020towards}
R.~Wang, K.~Kashinath, M.~Mustafa, A.~Albert, and R.~Yu, ``Towards
  physics-informed deep learning for turbulent flow prediction,'' in
  \emph{Proceedings of the 26th ACM SIGKDD International Conference on
  Knowledge Discovery \& Data Mining}, 2020, pp. 1457--1466.

\bibitem{xue2020amortized}
T.~Xue, A.~Beatson, S.~Adriaenssens, and R.~Adams, ``Amortized finite element
  analysis for fast pde-constrained optimization,'' in \emph{International
  Conference on Machine Learning}.\hskip 1em plus 0.5em minus 0.4em\relax PMLR,
  2020, pp. 10\,638--10\,647.

\bibitem{sanchez2018graph}
A.~Sanchez-Gonzalez, N.~Heess, J.~T. Springenberg, J.~Merel, M.~Riedmiller,
  R.~Hadsell, and P.~Battaglia, ``Graph networks as learnable physics engines
  for inference and control,'' in \emph{International Conference on Machine
  Learning}.\hskip 1em plus 0.5em minus 0.4em\relax PMLR, 2018, pp. 4470--4479.

\bibitem{wu2017learning}
J.~Wu, E.~Lu, P.~Kohli, B.~Freeman, and J.~Tenenbaum, ``Learning to see physics
  via visual de-animation,'' \emph{Advances in Neural Information Processing
  Systems}, vol.~30, 2017.

\bibitem{lu2019deeponet}
L.~Lu, P.~Jin, and G.~E. Karniadakis, ``Deeponet: Learning nonlinear operators
  for identifying differential equations based on the universal approximation
  theorem of operators,'' \emph{arXiv preprint arXiv:1910.03193}, 2019.

\bibitem{sirignano2018dgm}
J.~Sirignano and K.~Spiliopoulos, ``Dgm: A deep learning algorithm for solving
  partial differential equations,'' \emph{Journal of computational physics},
  vol. 375, pp. 1339--1364, 2018.

\bibitem{han2018solving}
J.~Han, A.~Jentzen, and W.~E, ``Solving high-dimensional partial differential
  equations using deep learning,'' \emph{Proceedings of the National Academy of
  Sciences}, vol. 115, no.~34, pp. 8505--8510, 2018.

\bibitem{meng2020ppinn}
X.~Meng, Z.~Li, D.~Zhang, and G.~E. Karniadakis, ``Ppinn: Parareal
  physics-informed neural network for time-dependent pdes,'' \emph{Computer
  Methods in Applied Mechanics and Engineering}, vol. 370, p. 113250, 2020.

\bibitem{liu2022predicting}
X.-Y. Liu, H.~Sun, and J.-X. Wang, ``Predicting parametric spatiotemporal
  dynamics by multi-resolution pde structure-preserved deep learning,''
  \emph{arXiv preprint arXiv:2205.03990}, 2022.

\bibitem{rao2021hard}
C.~Rao, H.~Sun, and Y.~Liu, ``Hard encoding of physics for learning
  spatiotemporal dynamics,'' \emph{arXiv preprint arXiv:2105.00557}, 2021.

\bibitem{rao2022discovering}
C.~Rao, P.~Ren, Y.~Liu, and H.~Sun, ``Discovering nonlinear pdes from scarce
  data with physics-encoded learning,'' \emph{arXiv preprint arXiv:2201.12354},
  2022.

\bibitem{huang2022meta}
X.~Huang, Z.~Ye, H.~Liu, S.~Ji, Z.~Wang, K.~Yang, Y.~Li, M.~Wang, H.~Chu, F.~Yu
  \emph{et~al.}, ``Meta-auto-decoder for solving parametric partial
  differential equations,'' \emph{Advances in Neural Information Processing
  Systems}, vol.~35, pp. 23\,426--23\,438, 2022.

\bibitem{horn1981determining}
B.~K. Horn and B.~G. Schunck, ``Determining optical flow,'' \emph{Artificial
  intelligence}, vol.~17, no. 1-3, pp. 185--203, 1981.

\bibitem{zach2007duality}
C.~Zach, T.~Pock, and H.~Bischof, ``A duality based approach for realtime tv-l
  1 optical flow,'' in \emph{Pattern Recognition: 29th DAGM Symposium,
  Heidelberg, Germany, September 12-14, 2007. Proceedings 29}.\hskip 1em plus
  0.5em minus 0.4em\relax Springer, 2007, pp. 214--223.

\bibitem{brox2010large}
T.~Brox and J.~Malik, ``Large displacement optical flow: descriptor matching in
  variational motion estimation,'' \emph{IEEE transactions on pattern analysis
  and machine intelligence}, vol.~33, no.~3, pp. 500--513, 2010.

\bibitem{menze2015discrete}
M.~Menze, C.~Heipke, and A.~Geiger, ``Discrete optimization for optical flow,''
  in \emph{Pattern Recognition: 37th German Conference, GCPR 2015, Aachen,
  Germany, October 7-10, 2015, Proceedings 37}.\hskip 1em plus 0.5em minus
  0.4em\relax Springer, 2015, pp. 16--28.

\bibitem{chen2016full}
Q.~Chen and V.~Koltun, ``Full flow: Optical flow estimation by global
  optimization over regular grids,'' in \emph{Proceedings of the IEEE
  conference on computer vision and pattern recognition}, 2016, pp. 4706--4714.

\bibitem{teed2020raft}
Z.~Teed and J.~Deng, ``Raft: Recurrent all-pairs field transforms for optical
  flow,'' in \emph{Computer Vision--ECCV 2020: 16th European Conference,
  Glasgow, UK, August 23--28, 2020, Proceedings, Part II 16}.\hskip 1em plus
  0.5em minus 0.4em\relax Springer, 2020, pp. 402--419.

\bibitem{jiang2021learning}
S.~Jiang, D.~Campbell, Y.~Lu, H.~Li, and R.~Hartley, ``Learning to estimate
  hidden motions with global motion aggregation,'' in \emph{Proceedings of the
  IEEE/CVF International Conference on Computer Vision}, 2021, pp. 9772--9781.

\bibitem{razavi2019generating}
A.~Razavi, A.~Van~den Oord, and O.~Vinyals, ``Generating diverse high-fidelity
  images with vq-vae-2,'' \emph{Advances in neural information processing
  systems}, vol.~32, 2019.

\bibitem{walker2021predicting}
J.~Walker, A.~Razavi, and A.~v.~d. Oord, ``Predicting video with vqvae,''
  \emph{arXiv preprint arXiv:2103.01950}, 2021.

\bibitem{bocus2023streamlining}
M.~J. Bocus, X.~Wang, R.~Piechocki \emph{et~al.}, ``Streamlining multimodal
  data fusion in wireless communication and sensor networks,'' \emph{arXiv
  preprint arXiv:2302.12636}, 2023.

\bibitem{liu2021conditional}
X.~Liu, T.~Iqbal, J.~Zhao, Q.~Huang, M.~D. Plumbley, and W.~Wang, ``Conditional
  sound generation using neural discrete time-frequency representation
  learning,'' in \emph{2021 IEEE 31st International Workshop on Machine
  Learning for Signal Processing (MLSP)}.\hskip 1em plus 0.5em minus
  0.4em\relax IEEE, 2021, pp. 1--6.

\bibitem{zhao2018unsupervised}
T.~Zhao, K.~Lee, and M.~Eskenazi, ``Unsupervised discrete sentence
  representation learning for interpretable neural dialog generation,''
  \emph{arXiv preprint arXiv:1804.08069}, 2018.

\bibitem{oh2015action}
J.~Oh, X.~Guo, H.~Lee, R.~L. Lewis, and S.~Singh, ``Action-conditional video
  prediction using deep networks in atari games,'' \emph{Advances in neural
  information processing systems}, vol.~28, 2015.

\bibitem{mathieu2015deep}
M.~Mathieu, C.~Couprie, and Y.~LeCun, ``Deep multi-scale video prediction
  beyond mean square error,'' \emph{arXiv preprint arXiv:1511.05440}, 2015.

\bibitem{tulyakov2018mocogan}
S.~Tulyakov, M.-Y. Liu, X.~Yang, and J.~Kautz, ``Mocogan: Decomposing motion
  and content for video generation,'' in \emph{Proceedings of the IEEE
  conference on computer vision and pattern recognition}, 2018, pp. 1526--1535.

\bibitem{ranzato2014video}
M.~Ranzato, A.~Szlam, J.~Bruna, M.~Mathieu, R.~Collobert, and S.~Chopra,
  ``Video (language) modeling: a baseline for generative models of natural
  videos,'' \emph{arXiv preprint arXiv:1412.6604}, 2014.

\bibitem{srivastava2015unsupervised}
N.~Srivastava, E.~Mansimov, and R.~Salakhudinov, ``Unsupervised learning of
  video representations using lstms,'' in \emph{International conference on
  machine learning}.\hskip 1em plus 0.5em minus 0.4em\relax PMLR, 2015, pp.
  843--852.

\bibitem{villegas2017learning}
R.~Villegas, J.~Yang, Y.~Zou, S.~Sohn, X.~Lin, and H.~Lee, ``Learning to
  generate long-term future via hierarchical prediction,'' in
  \emph{international conference on machine learning}.\hskip 1em plus 0.5em
  minus 0.4em\relax PMLR, 2017, pp. 3560--3569.

\bibitem{villegas2018hierarchical}
R.~Villegas, D.~Erhan, H.~Lee \emph{et~al.}, ``Hierarchical long-term video
  prediction without supervision,'' in \emph{International Conference on
  Machine Learning}.\hskip 1em plus 0.5em minus 0.4em\relax PMLR, 2018, pp.
  6038--6046.

\bibitem{kim2019variational}
T.~Kim, S.~Ahn, and Y.~Bengio, ``Variational temporal abstraction,''
  \emph{Advances in Neural Information Processing Systems}, vol.~32, 2019.

\bibitem{weissenborn2019scaling}
D.~Weissenborn, O.~T{\"a}ckstr{\"o}m, and J.~Uszkoreit, ``Scaling
  autoregressive video models,'' \emph{arXiv preprint arXiv:1906.02634}, 2019.

\bibitem{kumar2019videoflow}
M.~Kumar, M.~Babaeizadeh, D.~Erhan, C.~Finn, S.~Levine, L.~Dinh, and D.~Kingma,
  ``Videoflow: A flow-based generative model for video,'' \emph{arXiv preprint
  arXiv:1903.01434}, vol.~2, no.~5, p.~3, 2019.

\bibitem{dosovitskiy2020image}
A.~Dosovitskiy, L.~Beyer, A.~Kolesnikov, D.~Weissenborn, X.~Zhai,
  T.~Unterthiner, M.~Dehghani, M.~Minderer, G.~Heigold, S.~Gelly \emph{et~al.},
  ``An image is worth 16x16 words: Transformers for image recognition at
  scale,'' \emph{arXiv preprint arXiv:2010.11929}, 2020.

\bibitem{bai2022rainformer}
C.~Bai, F.~Sun, J.~Zhang, Y.~Song, and S.~Chen, ``Rainformer: Features
  extraction balanced network for radar-based precipitation nowcasting,''
  \emph{IEEE Geoscience and Remote Sensing Letters}, vol.~19, pp. 1--5, 2022.

\bibitem{sun2020predicting}
J.~Sun, J.~Zhang, Q.~Li, X.~Yi, Y.~Liang, and Y.~Zheng, ``Predicting citywide
  crowd flows in irregular regions using multi-view graph convolutional
  networks,'' \emph{IEEE Transactions on Knowledge and Data Engineering},
  vol.~34, no.~5, pp. 2348--2359, 2020.

\bibitem{wang2020deep}
S.~Wang, J.~Cao, and S.~Y. Philip, ``Deep learning for spatio-temporal data
  mining: A survey,'' \emph{IEEE transactions on knowledge and data
  engineering}, vol.~34, no.~8, pp. 3681--3700, 2020.

\bibitem{jiang2021dl}
R.~Jiang, D.~Yin, Z.~Wang, Y.~Wang, J.~Deng, H.~Liu, Z.~Cai, J.~Deng, X.~Song,
  and R.~Shibasaki, ``Dl-traff: Survey and benchmark of deep learning models
  for urban traffic prediction,'' in \emph{Proceedings of the 30th ACM
  international conference on information \& knowledge management}, 2021, pp.
  4515--4525.

\bibitem{jiang2019censnet}
X.~Jiang, P.~Ji, and S.~Li, ``Censnet: Convolution with edge-node switching in
  graph neural networks.'' in \emph{IJCAI}, 2019, pp. 2656--2662.

\bibitem{wang2022a2djp}
K.~Wang, Z.~Zhou, X.~Wang, P.~Wang, Q.~Fang, and Y.~Wang, ``A2djp: A two
  graph-based component fused learning framework for urban anomaly distribution
  and duration joint-prediction,'' \emph{IEEE Transactions on Knowledge and
  Data Engineering}, 2022.

\bibitem{wang2017predrnn}
Y.~Wang, M.~Long, J.~Wang, Z.~Gao, and P.~S. Yu, ``Predrnn: Recurrent neural
  networks for predictive learning using spatiotemporal lstms,'' \emph{Advances
  in neural information processing systems}, vol.~30, 2017.

\bibitem{wang2019eidetic}
Y.~Wang, L.~Jiang, M.-H. Yang, L.-J. Li, M.~Long, and L.~Fei-Fei, ``Eidetic 3d
  lstm: A model for video prediction and beyond,'' in \emph{International
  conference on learning representations}.

\bibitem{gao2022simvp}
Z.~Gao, C.~Tan, L.~Wu, and S.~Z. Li, ``Simvp: Simpler yet better video
  prediction,'' in \emph{Proceedings of the IEEE/CVF Conference on Computer
  Vision and Pattern Recognition}, 2022, pp. 3170--3180.

\bibitem{yang2022learning}
T.-Y. Yang, J.~Rosca, K.~Narasimhan, and P.~J. Ramadge, ``Learning physics
  constrained dynamics using autoencoders,'' \emph{Advances in Neural
  Information Processing Systems}, vol.~35, pp. 17\,157--17\,172, 2022.

\bibitem{jaques2019physics}
M.~Jaques, M.~Burke, and T.~Hospedales, ``Physics-as-inverse-graphics:
  Unsupervised physical parameter estimation from video,'' \emph{arXiv preprint
  arXiv:1905.11169}, 2019.

\bibitem{yu2020efficient}
W.~Yu, Y.~Lu, S.~Easterbrook, and S.~Fidler, ``Efficient and
  information-preserving future frame prediction and beyond,'' 2020.

\bibitem{pervez2021stability}
A.~Pervez and E.~Gavves, ``Stability regularization for discrete representation
  learning,'' in \emph{International Conference on Learning Representations},
  2021.

\bibitem{liu2021cross}
A.~H. Liu, S.~Jin, C.-I.~J. Lai, A.~Rouditchenko, A.~Oliva, and J.~Glass,
  ``Cross-modal discrete representation learning,'' \emph{arXiv preprint
  arXiv:2106.05438}, 2021.

\bibitem{peng2021generating}
J.~Peng, D.~Liu, S.~Xu, and H.~Li, ``Generating diverse structure for image
  inpainting with hierarchical vq-vae,'' in \emph{Proceedings of the IEEE/CVF
  Conference on Computer Vision and Pattern Recognition}, 2021, pp.
  10\,775--10\,784.

\bibitem{yan2021videogpt}
W.~Yan, Y.~Zhang, P.~Abbeel, and A.~Srinivas, ``Videogpt: Video generation
  using vq-vae and transformers,'' \emph{arXiv preprint arXiv:2104.10157},
  2021.

\bibitem{zhang2020videogen}
Y.~Zhang, W.~Yan, P.~Abbeel, and A.~Srinivas, ``Videogen: generative modeling
  of videos using vq-vae and transformers,'' 2020.

\bibitem{garbacea2019low}
C.~G{\^a}rbacea, A.~van~den Oord, Y.~Li, F.~S. Lim, A.~Luebs, O.~Vinyals, and
  T.~C. Walters, ``Low bit-rate speech coding with vq-vae and a wavenet
  decoder,'' in \emph{ICASSP 2019-2019 IEEE International Conference on
  Acoustics, Speech and Signal Processing (ICASSP)}.\hskip 1em plus 0.5em minus
  0.4em\relax IEEE, 2019, pp. 735--739.

\bibitem{tjandra2020transformer}
A.~Tjandra, S.~Sakti, and S.~Nakamura, ``Transformer vq-vae for unsupervised
  unit discovery and speech synthesis: Zerospeech 2020 challenge,'' \emph{arXiv
  preprint arXiv:2005.11676}, 2020.

\bibitem{de2019deep}
E.~De~B{\'e}zenac, A.~Pajot, and P.~Gallinari, ``Deep learning for physical
  processes: Incorporating prior scientific knowledge,'' \emph{Journal of
  Statistical Mechanics: Theory and Experiment}, vol. 2019, no.~12, p. 124009,
  2019.

\bibitem{brox2004high}
T.~Brox, A.~Bruhn, N.~Papenberg, and J.~Weickert, ``High accuracy optical flow
  estimation based on a theory for warping,'' in \emph{Computer Vision-ECCV
  2004: 8th European Conference on Computer Vision, Prague, Czech Republic, May
  11-14, 2004. Proceedings, Part IV 8}.\hskip 1em plus 0.5em minus 0.4em\relax
  Springer, 2004, pp. 25--36.

\bibitem{bruhn2005lucas}
A.~Bruhn, J.~Weickert, and C.~Schn{\"o}rr, ``Lucas/kanade meets horn/schunck:
  Combining local and global optic flow methods,'' \emph{International journal
  of computer vision}, vol.~61, pp. 211--231, 2005.

\bibitem{dosovitskiy2015flownet}
A.~Dosovitskiy, P.~Fischer, E.~Ilg, P.~Hausser, C.~Hazirbas, V.~Golkov, P.~Van
  Der~Smagt, D.~Cremers, and T.~Brox, ``Flownet: Learning optical flow with
  convolutional networks,'' in \emph{Proceedings of the IEEE international
  conference on computer vision}, 2015, pp. 2758--2766.

\bibitem{fischer2015flownet}
P.~Fischer, A.~Dosovitskiy, E.~Ilg, P.~H{\"a}usser, C.~Haz{\i}rba{\c{s}},
  V.~Golkov, P.~Van~der Smagt, D.~Cremers, and T.~Brox, ``Flownet: Learning
  optical flow with convolutional networks,'' \emph{arXiv preprint
  arXiv:1504.06852}, 2015.

\bibitem{ilg2017flownet}
E.~Ilg, N.~Mayer, T.~Saikia, M.~Keuper, A.~Dosovitskiy, and T.~Brox, ``Flownet
  2.0: Evolution of optical flow estimation with deep networks,'' in
  \emph{Proceedings of the IEEE conference on computer vision and pattern
  recognition}, 2017, pp. 2462--2470.

\bibitem{wu2021motionrnn}
H.~Wu, Z.~Yao, J.~Wang, and M.~Long, ``Motionrnn: A flexible model for video
  prediction with spacetime-varying motions,'' in \emph{Proceedings of the
  IEEE/CVF conference on computer vision and pattern recognition}, 2021, pp.
  15\,435--15\,444.

\bibitem{veillette2020sevir}
M.~Veillette, S.~Samsi, and C.~Mattioli, ``Sevir: A storm event imagery dataset
  for deep learning applications in radar and satellite meteorology,''
  \emph{Advances in Neural Information Processing Systems}, vol.~33, pp.
  22\,009--22\,019, 2020.

\bibitem{li2020fourier}
Z.~Li, N.~Kovachki, K.~Azizzadenesheli, B.~Liu, K.~Bhattacharya, A.~Stuart, and
  A.~Anandkumar, ``Fourier neural operator for parametric partial differential
  equations,'' \emph{arXiv preprint arXiv:2010.08895}, 2020.

\bibitem{guibas2021adaptive}
J.~Guibas, M.~Mardani, Z.~Li, A.~Tao, A.~Anandkumar, and B.~Catanzaro,
  ``Adaptive fourier neural operators: Efficient token mixers for
  transformers,'' \emph{arXiv preprint arXiv:2111.13587}, 2021.

\bibitem{wang2019memory}
Y.~Wang, J.~Zhang, H.~Zhu, M.~Long, J.~Wang, and P.~S. Yu, ``Memory in memory:
  A predictive neural network for learning higher-order non-stationarity from
  spatiotemporal dynamics,'' in \emph{Proceedings of the IEEE/CVF conference on
  computer vision and pattern recognition}, 2019, pp. 9154--9162.

\bibitem{wu2023pastnet}
H.~Wu, W.~Xion, F.~Xu, X.~Luo, C.~Chen, X.-S. Hua, and H.~Wang, ``Pastnet:
  Introducing physical inductive biases for spatio-temporal video prediction,''
  \emph{arXiv preprint arXiv:2305.11421}, 2023.

\end{thebibliography}
}

\begin{IEEEbiography}[{\includegraphics[width=1in,height=1.25in,clip,keepaspectratio]{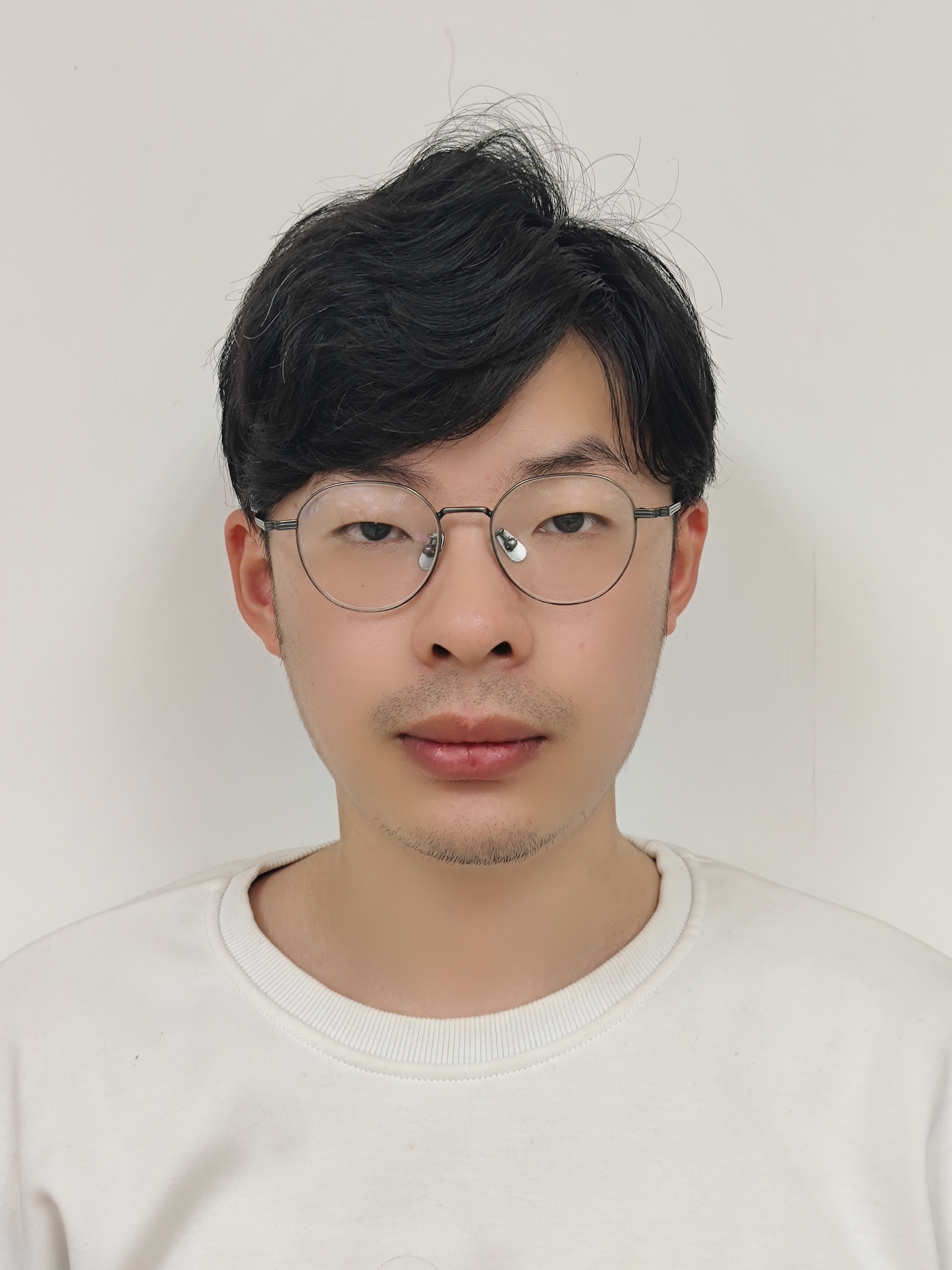}}]{Kun Wang} is currently a Ph.D. candidate of University of Science and Technology of China (USTC). He has published papers on top conferences and journals such as ICLR and TKDE. His primary research interests are Sparse Neural Networks, ML4Science and Graph Neural Networks.
\end{IEEEbiography}
%\vspace{-2.5in}
\vspace{-1.2cm}

\begin{IEEEbiography}[{\includegraphics[width=1in,height=1.25in,clip,keepaspectratio]{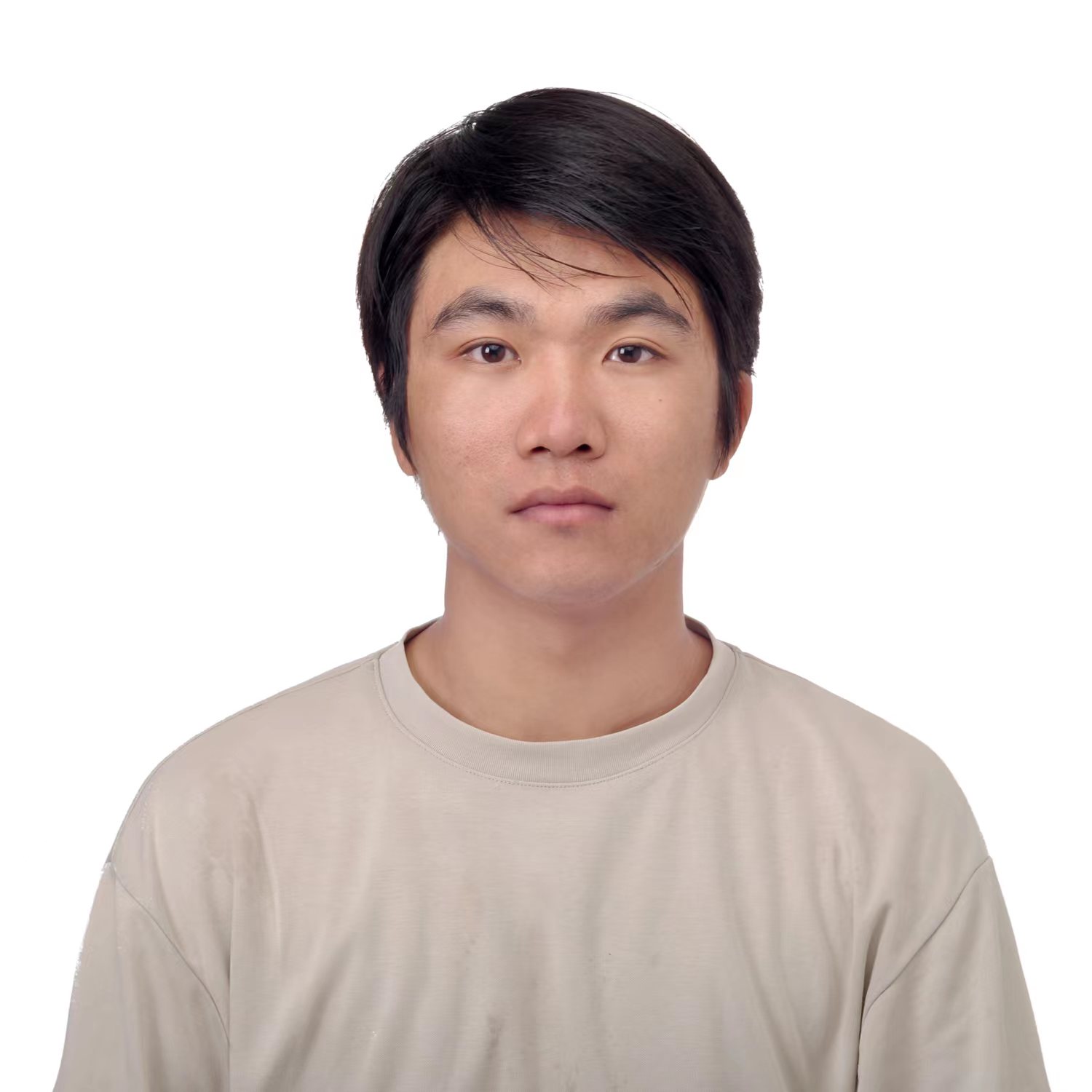}}]{Hao Wu} is a master's student who is currently undergoing joint training by the Department of Computer Science and Technology at the University of Science and Technology of China (USTC) and the Machine Learning Platform Department at Tencent TEG. His research interests encompass various areas, including spatio-temporal data mining, modeling of physical dynamical systems, and meta-learning, among others.
\end{IEEEbiography}
\vspace{-1.2cm}

\begin{IEEEbiography}[{\includegraphics[width=1in,height=1.25in,clip,keepaspectratio]{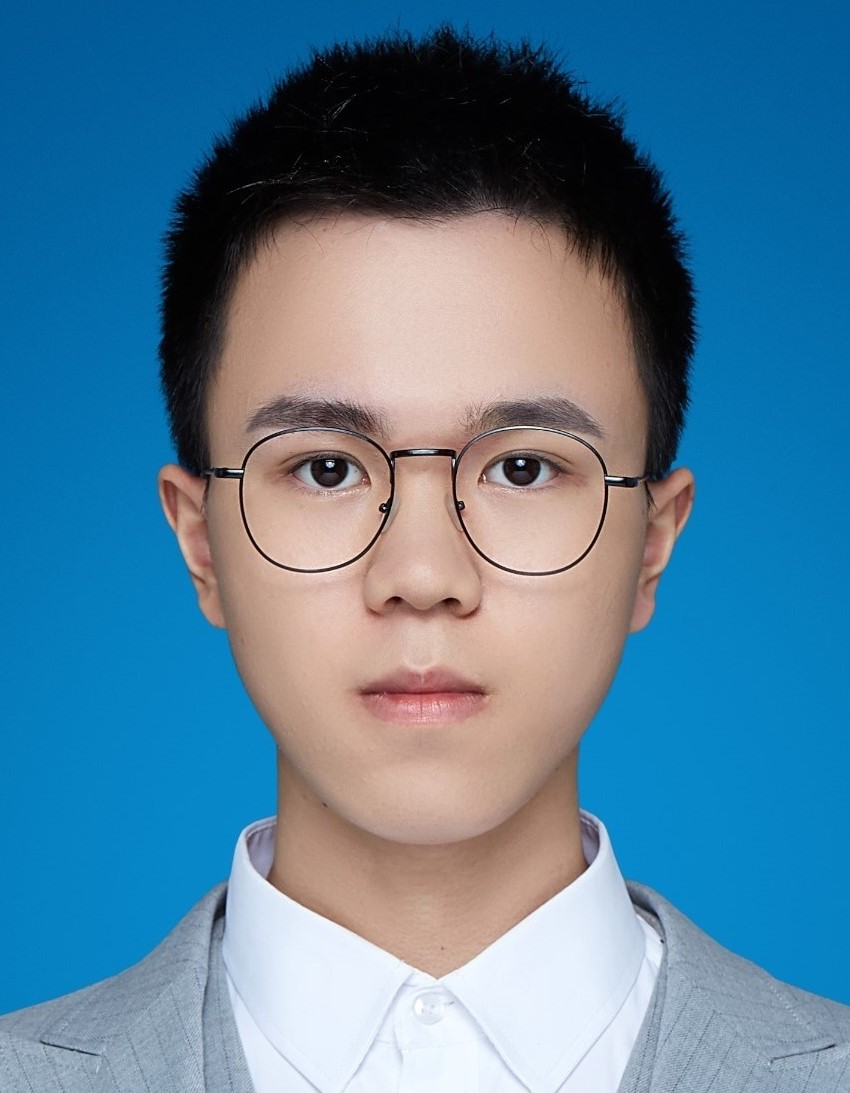}}]{Guibin Zhang} is currently an undergraduate in the Department of Computer Science and Technology, College of Electronic and Information Engineering, major in Data Science, Tongji University, Shanghai, China. His research interest include data mining, graph representation learning and knowledge modeling.
\end{IEEEbiography}
\vspace{-1.2cm}

\begin{IEEEbiography}[{\includegraphics[width=1in,height=1.25in,clip,keepaspectratio]{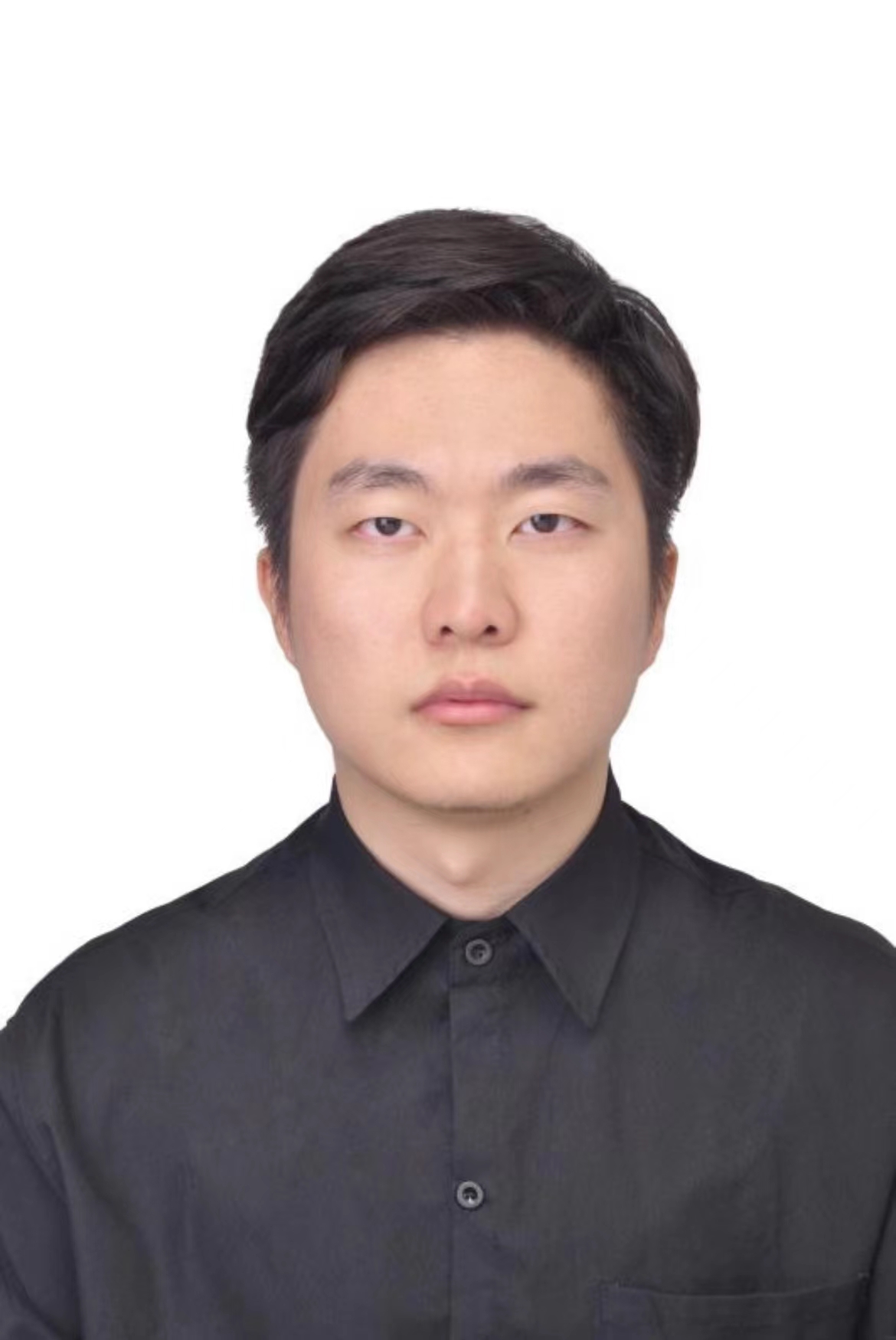}}]{Junfeng Fang} is currently a Ph.D.candidate of University of Science and Technology of China(USTC). His primary research interests are Trustable AI, Graph Neural Network and Diffusion Model.
\end{IEEEbiography}
\vspace{-1.2cm}

\begin{IEEEbiography}[{\includegraphics[width=1in,height=1.25in,clip,keepaspectratio]{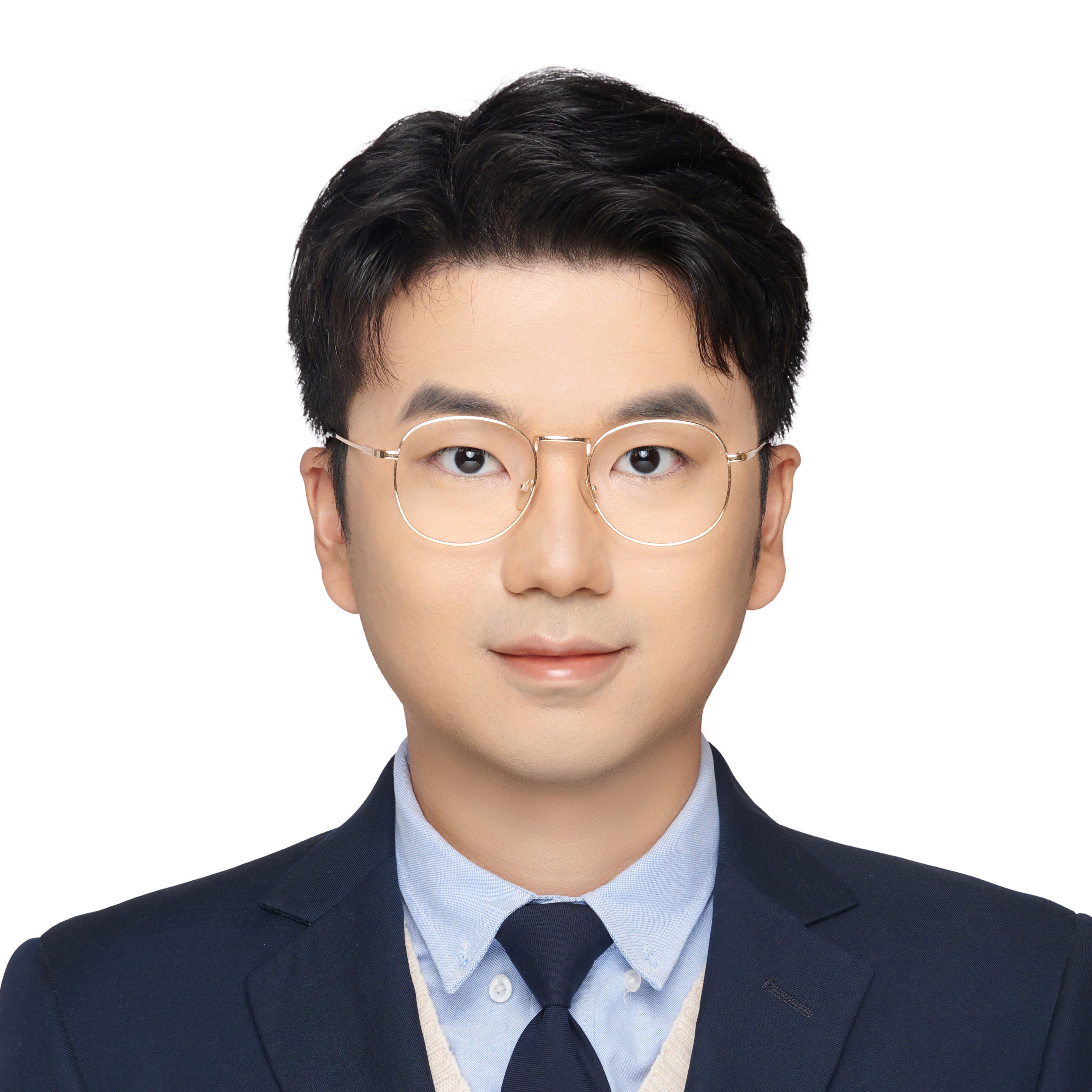}}]{Yuxuan Liang} (Member, IEEE) is currently an Assistant Professor at Intelligent Transportation Thrust, also affiliated with Data Science and Analytics Thrust, Hong Kong University of Science and Technology (Guangzhou). He is working on the research, development, and innovation of spatio-temporal data mining and AI, with a broad range of applications in smart cities. Prior to that, he completed his PhD study at NUS. He published over 40 peer-reviewed papers in refereed journals and conferences, such as TKDE, AI Journal, TMC, KDD, WWW, NeurIPS, and ICLR. He was recognized as 1 out of 10 most innovative and impactful PhD students focusing on data science in Singapore by Singapore Data Science Consortium.
\end{IEEEbiography}
\vspace{-1.2cm}

\begin{IEEEbiography}[{\includegraphics[width=1in,height=1.25in,clip,keepaspectratio]{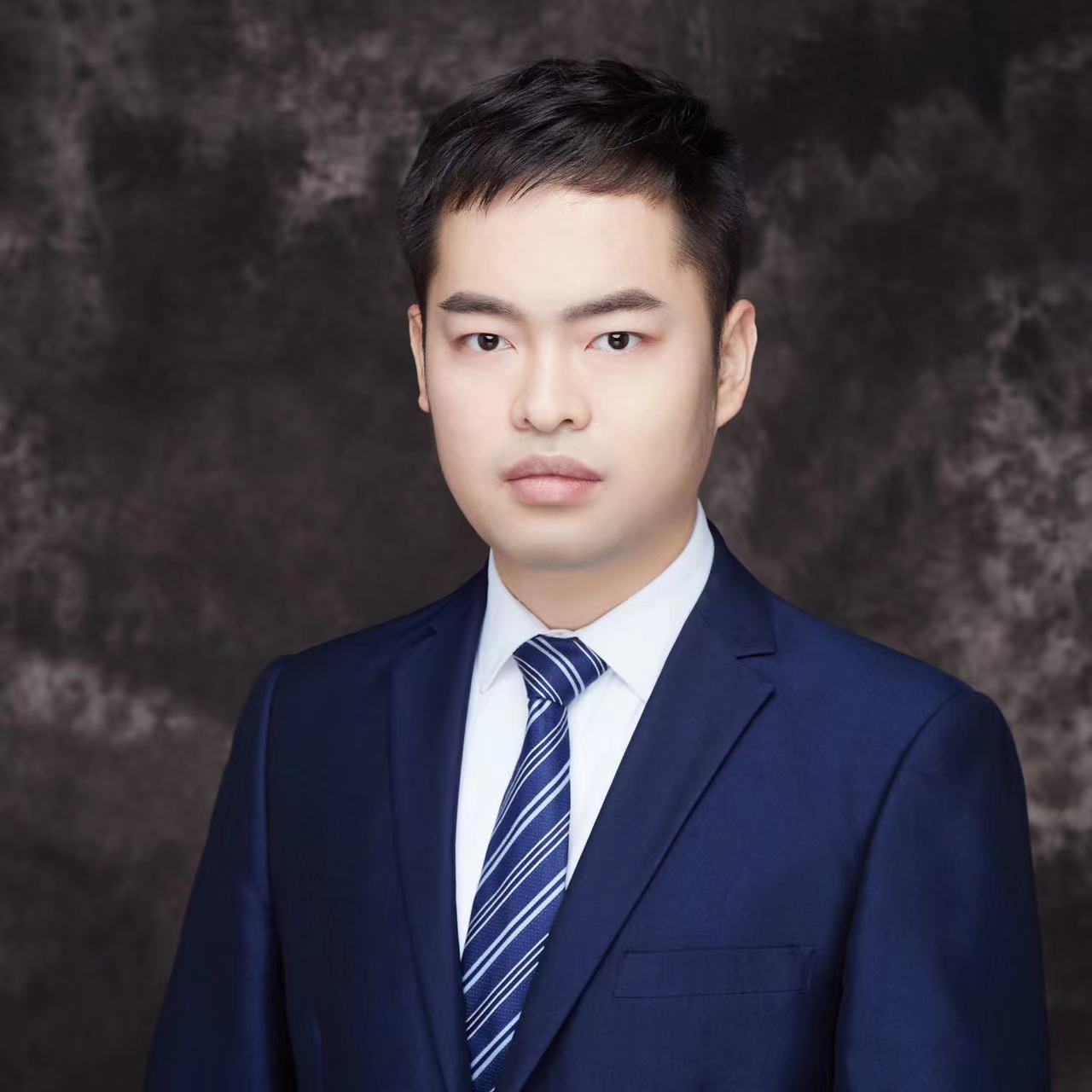}}]{Yuankai Wu} (Member, IEEE) received the Ph.D. degree from the Beijing Institute of Technology (BIT), Beijing, China, in  2019. He is a professor at the College of Computer Science, Sichuan University, China. Prior to joining Sichuan University in March 2022, he was an IVADO postdoc researcher with the Department of Civil Engineering, McGill University. His research interests include spatiotemporal data analysis, intelligent transportation systems, and intelligent decision-making.
\end{IEEEbiography}
\vspace{-1.2cm}

\begin{IEEEbiography}
[{\includegraphics[width=1in,height=1.25in,clip,keepaspectratio]{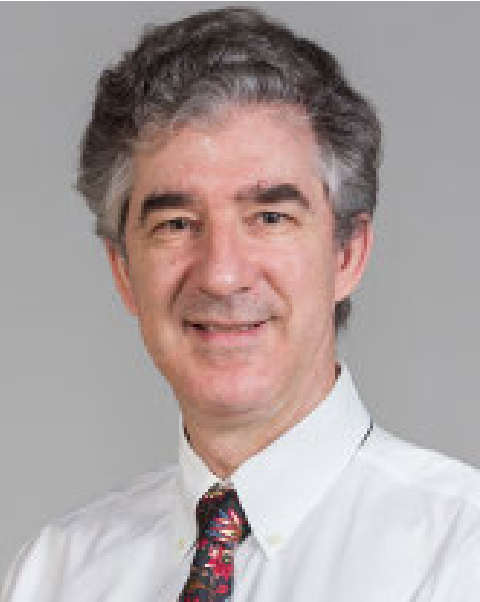}}]{Roger Zimmermann} (Senior Member, IEEE) received the M.S. and Ph.D. degrees from the University of Southern California (USC), in 1994 and 1998, respectively. He is a Professor with the Department of Computer Science, National University of Singapore (NUS). He is also the Deputy Director with the Smart Systems Institute (SSI) and a Key Investigator with the Grab-NUS AI Lab. He has coauthored a book, seven patents, and more than 300 conference
publications, journal articles, and book chapters. His research interests include streaming media architectures, multimedia networking, applications of machine/deep learning, and spatial data analytics. He is currently an Associate Editor for IEEE MultiMedia, Transactions on Multimedia Computing, Communications, and Applications (TOMM) (ACM), Multimedia Tools and Applications (MTAP) (Springer), and the IEEE Open Journal of the Communications Society (OJ-COMS). He is a distinguished member of the ACM and a senior member of the IEEE.
\end{IEEEbiography}

\begin{IEEEbiography}[{\includegraphics[width=1in,height=1.25in,clip,keepaspectratio]{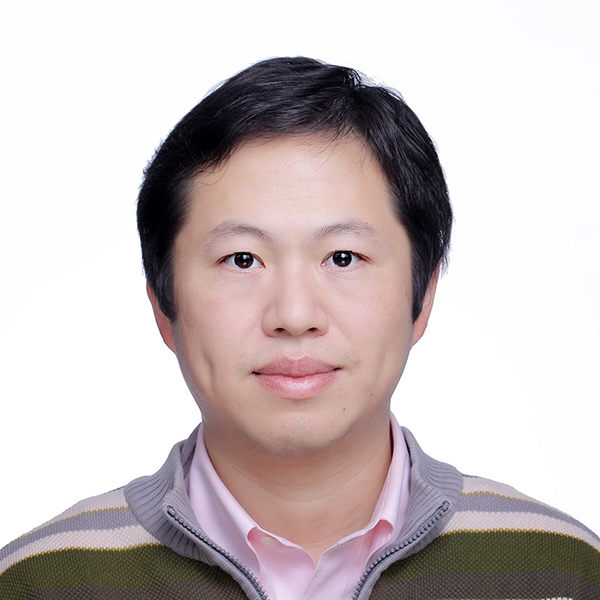}}]{Yang Wang} is now an associate professor at USTC. He got his Ph.D. degree at University of Science and Technology of China in 2007, under supervision of Professor Liusheng Huang. He also worked as a postdoc at USTC with Professor Liusheng Huang. His research interest mainly includes wireless (sensor) networks, distributed systems, data mining, and machine learning. 
\end{IEEEbiography}
\vspace{-1.2cm}

% \appendices

% % you can choose not to have a title for an appendix
% % if you want by leaving the argument blank
% \section{}

% \ifCLASSOPTIONcompsoc
%   % The Computer Society usually uses the plural form
%   \section*{Acknowledgments}
% \else
%   % regular IEEE prefers the singular form
% \fi


\ifCLASSOPTIONcaptionsoff
  \newpage
\fi

\end{document}